\documentclass[
]{jss}

\usepackage{orcidlink,thumbpdf,lmodern}

\usepackage[utf8]{inputenc}

\author{
Patrick Schratz~\orcidlink{0000-0003-0748-6624}\\Friedrich Schiller
University Jena \And Marc
Becker~\orcidlink{0000-0002-8115-0400}\\Ludwig-Maximilians-Universität
München \AND Michel Lang~\orcidlink{0000-0001-9754-0393}\\TU
Dortmund University \And Alexander
Brenning~\orcidlink{0000-0001-6640-679X}\\Friedrich Schiller University
Jena
}
\title{\pkg{mlr3spatiotempcv}: Spatiotemporal resampling methods for
machine learning in R}

\Plainauthor{Patrick Schratz, Marc Becker, Michel Lang, Alexander
Brenning}
\Plaintitle{mlr3spatiotempcv: spatiotemporal resampling methods for
machine learning in R}

\Abstract{
Spatial and spatiotemporal machine-learning models require a suitable
framework for their model assessment, model selection, and
hyperparameter tuning, in order to avoid error estimation bias and
over-fitting. This contribution reviews the state-of-the-art in spatial
and spatiotemporal cross-validation, and introduces the \proglang{R}
package \pkg{mlr3spatiotempcv} as an extension package of the
machine-learning framework \pkg{mlr3}. Currently various \proglang{R}
packages implementing different spatiotemporal partitioning strategies
exist: \pkg{blockCV}, \pkg{CAST}, \pkg{skmeans} and \pkg{sperrorest}.
The goal of \pkg{mlr3spatiotempcv} is to gather the available
spatiotemporal resampling methods in \proglang{R} and make them
available to users through a simple and common interface. This is made
possible by integrating the package directly into the \pkg{mlr3}
machine-learning framework, which already has support for generic
non-spatiotemporal resampling methods such as random partitioning. One
advantage is the use of a consistent nomenclature in an overarching
machine-learning toolkit instead of a varying package-specific syntax,
making it easier for users to choose from a variety of spatiotemporal
resampling methods. This package avoids giving recommendations which
method to use in practice as this decision depends on the predictive
task at hand, the autocorrelation within the data, and the spatial
structure of the sampling design or geographic objects being studied.
}

\Keywords{cross-validation, predictive performance, machine
learning, autocorrelation, spatial, temporal, \proglang{R}}
\Plainkeywords{cross-validation, predictive
performance, autocorrelation, spatial, temporal, R}

\Address{
    Patrick Schratz\\
    Friedrich Schiller University Jena\\
    Department of Geography\\
Geographic Information Science group\\
  E-mail: \email{patrick.schratz@uni-jena.de}\\

      Marc Becker\\
    Ludwig-Maximilians-Universität München\\
    Department of Statistics\\
Statistical Learning and Data Science group\\
  E-mail: \email{marc.becker@stat.uni-muenchen.de}\\

      Michel Lang\\
    TU Dortmund University\\
    Faculty of Statistics\\
  E-mail: \email{lang@statistik.tu-dortmund.de}\\

      Alexander Brenning\\
    Friedrich Schiller University Jena\\
    Department of Geography\\
Geographic Information Science group\\
  E-mail: \email{alexander.brenning@uni-jena.de}\\

  }

\providecommand{\tightlist}{%
  \setlength{\itemsep}{0pt}\setlength{\parskip}{0pt}}

\usepackage{amsmath} \usepackage{amssymb} \usepackage{booktabs} \usepackage{longtable} \usepackage{adjustbox} \usepackage{multirow} \usepackage{tabularx} \usepackage{placeins} \usepackage{scrextend} \usepackage{underscore} \usepackage{tablefootnote} 

\begin{document}

\nocite{ploton2020}

\hypertarget{sec:intro}{%
\section{Introduction}\label{sec:intro}}

Spatial and spatiotemporal prediction tasks are common in applications
ranging from environmental sciences to archaeology and epidemiology.
While sophisticated mathematical frameworks have long been developed in
spatial statistics to characterize predictive uncertainties under
well-defined mathematical assumptions such as intrinsic stationarity
\citep[e.g.,][]{cressie1993}, computational estimation procedures have
only been proposed more recently to assess predictive performances of
spatial and spatiotemporal prediction models
\citep{brenning2005, brenning2012, pohjankukka2017, roberts2017}.

Although alternatives such as the bootstrap exist since some decades
\citep{efron1983, hand1997}, cross-validation (CV) is a particularly
well-established, easy-to-implement algorithm for \emph{model
assessment} of supervised machine-learning models \citep[ and next
section]{efron1983} and \emph{model selection} \citep{arlot2010}. In its
basic form, CV is based on resampling the data without paying attention
to any possible dependence structure, which may arise from, e.g.,
grouped or structured data, or underlying environmental processes
inducing some sort of spatial coherence at the landscape scale. In
treating dependent observations as independent, or ignoring
autocorrelation, CV test samples may in fact be heavily correlated with,
or even pseudo-replicates of, the data used for training the model,
which introduces a potentially severe bias in assessing the
transferability of flexible machine-learning (ML) models.

This CV bias is well-known in spatial as well as non-spatial prediction
\citep{brenning2005, brenning2008, arlot2010, roberts2017} and in
forecasting \citep{bergmeir2018}. It is most easily understood from a
predictive modeling perspective by focusing on the question of where
(and when) the model should be used for prediction. In crop
classification from remotely-sensed data, for instance, learning samples
routinely contain multiple grid cells from a sample of fields with known
crop type, for instance 2000 grid cells from 100 fields scattered across
a large study region. The purpose of training a model on this particular
sample is to make predictions on other, new fields within the same
geographic domain \citep[\emph{intra-domain}
prediction,][]{brenning2005} --- not \emph{within} the same field, which
obviously presents only a single crop type that is already known from
the training sample. In this specific situation it would therefore seem
rather unwise to train a model on a simple random subsample of grid
cells, and to test it on the remaining data, using other grid cells from
the same fields, as if one wanted to predict within a field. The results
from this performance assessment would be over-optimistic, and perhaps
badly so. To mimic the predictive situation for which the model is
trained, one would rather have to resample at the level of fields, not
grid cells \citep{pena2015}. If the model was to be applied to adjacent
agricultural regions, i.e., outside the learning sample's spatial domain
\citep[\emph{extra-domain} prediction,][]{brenning2005}, it would even
seem necessary to resample at a higher level of spatial aggregation,
i.e., at the level of sub-regions within the learning sample, in order
to realistically mimic the actual prediction task. The CV resampling
needed therefore depends as much on the prediction task itself as on the
data structure or dependency at hand.

While it is not the purpose of this article to recommend specific
resampling schemes for specific use cases, the example from above may
suffice to motivate the use of appropriate spatial and spatiotemporal
cross-validation techniques, and the need for a unified framework and
computational toolbox that accommodate a variety of prediction tasks
that may be applicable to a broad range of application scenarios.
\pkg{mlr3spatiotempcv} is such a toolbox.

This toolbox, implemented as an open-source R package, builds upon and
generalizes several existing toolboxes that have been developed in
recent years for more specific settings (Table \ref{tab:sptcv-methods}).
The earliest and most comprehensive of these implementations is the
\pkg{sperrorest} \proglang{R} package \citep{brenning2012}, which
provides an extensible framework and includes predefined resampling
strategies based on geometric blocking, clustering, and buffering. In
contrast, packages \pkg{blockCV} and \pkg{ENMeval} were developed for
block and buffer resampling with a focus on species distribution
modeling \citep{blockCV, rest2014, muscarella2014}. However, neither of
these have been integrated into established machine-learning frameworks
such as \pkg{mlr}/\pkg{mlr3} \citep{mlr3} or
\pkg{caret}/\pkg{tidymodels} \citep{kuhn2020}, and all of them lack
support for temporal prediction tasks. The \pkg{CAST} package, in
contrast, focuses on spatiotemporal prediction tasks and makes use of
some functions of the \pkg{caret} framework \citep{cast, meyer2018}. One
limitation of all these packages is the sole focus on model assessment,
while the proposed implementation within the \pkg{mlr3} framework also
offers seamless integration into model selection and provides parallel
execution and enhanced logging abilities. It is worth noting that a
spatial cross-validation library named \pkg{spacv} has recently been
developed for Python3, which can be used with the \pkg{scikit-learn}
machine-learning framework \citep{pedregosa2011}.

Thus, \pkg{mlr3spatiotempcv} implements for the first time a
comprehensive state-of-the-art compilation of spatial and spatiotemporal
partitioning schemes that is well-integrated into a comprehensive
machine-learning framework in R, the \pkg{mlr3} ecosystem. This package
is furthermore equipped with a variety of two- and three-dimensional
visualization capabilities. The hope is that this implementation will
simplify and facilitate reproducible geospatial modeling and
code-sharing across a broad range of application domains.

The purpose of this article is to give an overview of the methods
implemented in the \proglang{R} package \pkg{mlr3spatiotempcv}. After
presenting the conceptual background in the following section, the
overall structure of the \pkg{mlr3spatiotempcv} package is outlined.
Next, various spatial and spatiotemporal partitioning techniques are
contrasted and compared, before their application is demonstrated in a
machine-learning model assessment in the following section. Finally,
recommendations for the selection of suitable resampling techniques are
given.

\hypertarget{spatial-and-spatiotemporal-cv}{%
\section{Spatial and spatiotemporal
CV}\label{spatial-and-spatiotemporal-cv}}

In CV for predictive model assessment, the following formal setting is
considered. The interest is in predicting a numerical or categorical
response \(y\) of an object or instance using a feature vector
\(\mathbf{x} = (x^{(1)}, \ldots, x^{(p)})^t\in\mathbb{R}^p\) and a model
\(\hat{f}_\mathcal{L}\) that has been trained on a learning sample
\(\mathcal{L} = \{(y_i, \mathbf{x}_i),\ i = 1, ..., n\}\). The goal is
to estimate the expected value of the performance of
\(\hat{f}_\mathcal{L}\), \[
\mathit{perf(\hat{f}_\mathcal{L})} := E(l(Y,\hat{f}_\mathcal{L}(X))),
\] where \(l\) is a real-valued loss function, and the expected value is
with respect to the probability distribution of \(X\), the features of
an instance \((Y,X)\) drawn randomly from the underlying population.
This is referred to as the \emph{actual} or \emph{conditional}
performance measure, as it is conditional on \(\mathcal{L}\)
\citep{hand1997}. The loss function can take a variety of forms such as
the misclassification error \(I(Y\neq\hat{f}_\mathcal{L}(X))\) in
classification, or the squared error \((Y-\hat{f}_\mathcal{L}(X))^2\) in
regression, among many other possible measures. The choice of the
performance measure is equally critical as the choice of the estimation
procedure, but it is beyond the scope of this contribution to discuss
performance measures for regression and classification (see, e.g.,
\citet{hand1997} for classification, and \citet{hyndman2006} for
regression and forecasting tasks).

Since there is only a sample \(\mathcal{T}\) of test data drawn from the
population, one can only \emph{estimate} the conditional performance of
\(\hat{f}_\mathcal{L}\): \[
\widehat{\mathit{perf}}_T(\hat{f}_\mathcal{L}) = \frac{1}{|\mathcal{T}|}\sum_{(Y,X)\in\mathcal{T}}l(Y,\hat{f}_\mathcal{L}(X)).
\] This representation as a point estimator of
\(\mathit{perf(\hat{f}_\mathcal{L})}\) underlines the importance of
using a random sample for model assessment to avoid estimation bias.
Other estimators than the simple mean may be required when
\(\mathcal{T}\) is not a simple random sample, for instance a stratified
random sample \citep[e.g.,][]{thompson2012}. As always, judgment
sampling may lead to uncontrollable bias.

Since re-using the learning sample \(\mathcal{L}\) for testing, i.e.,
\(\mathcal{T}:=\mathcal{L}\), would yield the over-optimistic
\emph{resubstitution} or \emph{apparent} performance, CV partitions the
sample \(\mathcal{L}\) into disjoint training and test sets.
Specifically, \(\mathcal{L}\) is split into \(k\) partitions, \[
\mathcal{L} = \mathcal{L}_1 \cup \ldots \cup \mathcal{L}_k,\qquad \mathcal{L}_i\cap \mathcal{L}_j = \emptyset\quad \textrm{for all}\ i\neq j,
\] and a model \(\hat{f}_{(i)}\) is fitted on
\(\mathcal{L}_{(i)} := \mathcal{L}\setminus \mathcal{L}_i\), while
\(\mathcal{L}_i\) is withheld for testing. This is repeated for
\(i=1,\ldots,k\) in order to effectively use the entire sample for
testing, while keeping training and test sets disjoint at all times. The
\(k\)-fold CV estimator can therefore be written as \[
\widehat{\mathit{perf}}_{\mathcal{L}, CV}(f) := \frac{1}{k}\sum_{i=1}^k\widehat{\mathit{perf}}_{\mathcal{L}_i}(\hat{f}_{\mathcal{L}_{(i)}}),
\] where \(f\) is a ML algorithm, i.e., a mapping that trains a model
\(\hat{f}_\mathcal{S}\) using any suitable training sample
\(\mathcal{S}\). The use of \(k=5\) or \(k=10\) folds is most commonly
seen in practice, and these preferences are also supported by theory
\citep{bengio2004, cawley2010}. The \(k\)-fold CV estimator of model
performance is a nearly unbiased estimator of the conditional
performance measure when the observations were drawn independently
\citep{efron1983}. Since
\(\widehat{\mathit{perf}}_{\mathcal{L}, CV}(f)\) still depends on the
particular partitioning chosen for \(\mathcal{L}\), it is sometimes
recommended to repeat the estimation using different random
partitionings (\(r\)-repeated \(k\)-fold cross-validation) to reduce the
influence of randomness when creating partitions
\citep{vanwinckelen2012}.

In traditional CV, the partitioning is based on uniform random
resampling, which ignores spatial or temporal autocorrelation or any
existing grouping structure as well as the structure of the prediction
task, and may result in over-optimistic performance estimates. Several
approaches have therefore been proposed in the literature and
implemented in software to accommodate a variety of predictive
situations (Table \ref{tab:sptcv-methods}).

Approaches based on \emph{spatial blocking} (or sometimes called
\emph{grouping}) require either the construction of spatial zones, or
the use of pre-existing spatial structures in the data. Let's refer to
these spatial units or blocks as \(\mathcal{Z}_i\), \(1\le i\le n_z\).
These blocks are often constructed to serve as the \(k=n_z\) spatial
partitions, for example by performing \(k\)-means clustering of the
sample coordinates \citep{russ2010}, which we refer to as
\emph{coordinate-based clustering}; or generating the desired number of
rectangular blocks as an example of \emph{geometric partitioning}. The
blocks may also be defined by a modeler based on an arbitrary
partitioning of the study region based on an external data source, which
we refer to as \emph{custom resampling}. This often used when the data
is grouped. For example, when using to multi-level sampling designs or
studying spatial objects, it has been proposed to apply LOO at the site
level \citep{martin2008, kasurak2011} or, in animal movement studies, at
the animal level \citep{anderson2005}. We will broadly refer to such
groups of observations as `blocks' in a generic sense, regardless of the
shape or origin of the groups. Also, data can be partitioned in feature
space instead of geographic space, which has been referred to as
``environmental blocking'' \citep{roberts2017}.

When \(n_z\) is much larger than the desired number of folds, \(k\),
then a partitioning can be applied to the zones themselves. In this
case, the zone indices \(1, \ldots, n_z\) are grouped into \(k\) equally
sized subsets \(\mathcal{I}_1, \ldots, \mathcal{I}_k\). This approach
has been applied, for example, in spatial CV at the agricultural field
level \citep{pena2015}. We would like to emphasize the conceptual
distinction between \emph{CV at the block level}, referring to this
scenario, and \emph{leave-one-block-out CV}, where the blocks themselves
define the CV partitions. Figure \ref{fig:rsmp-schema} gives an overview
of the conceptual framework and terminology used in this work.

One variant of CV is leave-one-out (LOO) CV, which has long been
established in geostatistics \citep{cressie1993}, sometimes with a focus
on the spatial distribution of LOO error \citep{willmott2006}. Although
this is just a special case of non-spatial CV with \(k=n\), it is
sometimes also referred to as spatial CV \citep{willmott2006}.

Spatial variants of CV have been proposed that apply an exclusion
\emph{buffer} or guard zone to the test locations to separate them from
the training data \citep{brenning2005, roberts2017}. One approach that
has been proposed for defining a separation distance is to use the range
of autocorrelation of model residuals to determine the buffer distance,
as this seeks to establish independence conditional on the predictors
\citep{brenning2005, roberts2017}.

It should be noted that \(k\)-fold CV with a large value of \(k\), and
LOO CV in particular (\(k=n\)), is not only very time-consuming since
the model has to be trained \(k\) times; these models will also be
nearly identical since only a tiny fraction of the data is withheld, and
therefore estimation bias increases. `Pure' LOO CV is therefore not
recommended for machine-learning model assessment.

In the purely temporal domain, a special case is to leave out temporal
observational units (or time slices; leave-time-out or LTO CV), as in
leave-one-year-out CV \citep{anderson2005, brenning2005}. CV and
hold-out validation strategies for time series have been discussed more
extensively in the forecasting literature, considering also the effects
of serial autocorrelation \citep{bergmeir2018}; these methods are not
the focus of the implementation presented in this work.

Turning to prediction tasks with spatiotemporal data, various spatial,
temporal, or spatiotemporal partitioning strategies are being used,
depending on the specific study objectives. While the former two ignore
the temporal and spatial dimension of the data, respectively, it has
also been proposed to leave out random subsets of locations and time
points \citep{meyer2018} or spatiotemporal clusters \citep{cluto}.
Details of these and other implementations are outlined in the
respective subsections of Section \ref{sec:implementation}.

\section[mlr3spatiotempcv within the mlr3 ecosystem]{\pkg{mlr3spatiotempcv}
within the \pkg{mlr3} ecosystem}\label{r-code}

With the increased awareness of the importance of spatial and
spatiotemporal resampling strategies and the growing popularity of R in
environmental modeling and geocomputation, it is important to equip ML
frameworks such as \pkg{mlr3} with suitable algorithms. In this context,
the \pkg{mlr3} ecosystem stands out as a unified, object-oriented and
extensible framework designed to accommodate numerous ML tasks with a
variety of learners, feature and model selection tools, and model
assessment capabilities \citep{mlr3, mlr3book}. All of these are
supported by advanced visualization tools, which are particularly
important in a spatial and spatiotemporal setting. Additionally,
\pkg{mlr3pipelines} \citep{mlr3pipelines} provides a plethora of
preprocessing operators to conveniently build ML pipelines which can be
resampled, tuned and benchmarked as regular learners.

With its integrative approach and its aim to provide long-term support,
\pkg{mlr3} overcomes the challenges of combining multiple specialized
packages with poorly standardized interfaces. Issues that practitioners
often face include varying argument lists of learners, different return
values of \texttt{predict()} methods, and support for only specific
feature types. These challenges result in substantial overhead and
possible reproducibility issues, which are exacerbated by asynchronous
development timelines of different components of the used ML pipelines.

Within the \pkg{mlr3} ecosystem, partitioning strategies are represented
by their own objects of class \texttt{Resampling}, most of which are
available within \pkg{mlr3} itself (e.g., random CV); other specialized
strategies are defined in extension packages such as
\pkg{mlr3spatiotempcv}. In the ML pipeline, these objects define the
data splits used for model assessment and selection (hyperparameter
tuning) by ML algorithms. Spatial and spatiotemporal partitioning
techniques in \pkg{mlr3spatiotempcv} are currently mostly imported and
interfaced from other packages, in particular \pkg{sperrorest},
\pkg{blockCV} and \pkg{CAST} \citep{brenning2012, blockCV, cast}, in
order to expose them to \pkg{mlr3} functionality. To reduce
dependencies, some methods were re-implemented instead of importing them
from the respective upstream packages.

Resampling objects in \pkg{mlr3spatiotmpcv} inherit from class
\texttt{mlr3::Resampling} and can be created from established object
classes for geospatial data in \proglang{R}, including simple features
\citep{pebesma2018}, which facilitates their integration into
domain-specific workflows in the geospatial sciences. Support for
projected (planar) and unprojected (geographic) coordinate reference
systems (CRS) currently varies depending on the partitioning techniques
used, since these inherit their behavior from the underlying upstream
packages.

Partitioning objects in \pkg{mlr3spatiotempcv} are equipped with generic
\texttt{plot()} and \texttt{autoplot()} methods to visualize the created
partitions. \texttt{autoplot()} is \pkg{ggplot2}-based and uses
\href{https://ggplot2.tidyverse.org}{ggplot2} \citep{ggplot2} in
two-dimensional geographic space and
\href{https://github.com/ropensci/plotly}{\pkg{plotly}} \citep{plotly}
in the three dimensional case, i.e., geographic space plus time.

While \pkg{mlr3spatiotempcv} solely focuses on spatiotemporal resampling
methods and their visualization, other packages such as
\pkg{mlr3spatial} or \pkg{mlr3temporal} are planned in the \pkg{mlr3}
ecosystem to provide dedicated spatiotemporal learner and prediction
methods.

From a user perspective, this package structure results in the following
workflow for model assessment with \pkg{mlr3spatiotempcv} within
\pkg{mlr3}: After choosing a ML algorithm that is supported by
\pkg{mlr3} and setting up a learner object, users need to select
hyperparameters that should be tuned and specify these in a
\texttt{paradox::ParamSet}. Next, a suitable resampling scheme available
within \pkg{mlr3spatiotempcv} is selected that mimics the spatial and/or
temporal structure of the prediction task, such as spatial
extrapolation, or forecasting of spatial time series. This information
is used to create a \texttt{Resampling} object which is used within a
(nested) CV to estimate the model performance. When using nested CV, the
resampling schemes in the inner (tuning, \texttt{mlr3tuning::AutoTuner})
and outer loop (performance estimation, \texttt{resample()}) should be
identical \citep{schratz2019}. To evaluate the (nested) resampling, an
adequate performance measure with respect to the response variable, such
as the misclassification rate (classification) or the root-mean-square
error (regression), must be selected and specified within
\texttt{mlr3tuning::AutoTuner} and \texttt{resample\$score()}. These
choices now allow the user to execute the model assessment via either
\texttt{resample()} (single model) or \texttt{benchmark()} (multiple
models), and the results can be summarized visually (via \pkg{mlr3viz})
or in tabular form by accessing the respective fields of the returned
\texttt{ResampleResult} object.

Additional examples and tutorial can be found in the
\href{https://mlr3book.mlr-org.com}{mlr3book} or the
\href{https://mlr-org.com/gallery}{mlr3gallery}.

\hypertarget{sec:implementation}{%
\section{Spatiotemporal partitioning methods and their
implementation}\label{sec:implementation}}

At the most general level, resampling methods are categorized according
to the level at which the data is partitioned and resampled (see Figure
\ref{fig:rsmp-schema}):

\begin{itemize}
\tightlist
\item
  \textbf{Spatial leave-one-out resampling}: Each individual observation
  forms a test set;
\item
  \textbf{Leave-one-block-out CV}: Individual blocks are left out as
  test data, i.e., the number of folds equals the number of blocks;
\item
  \textbf{CV at the block level}: Blocks are grouped into \(k\)
  partitions, each of which is used as a test fold.
\end{itemize}

In this context, a block can refer to an arbitrarily shaped spatial (or
spatiotemporal) group of observations, not necessarily a rectangular
region. A finer distinction can then be made by looking at how the
blocks are derived:

\begin{itemize}
\tightlist
\item
  Using a geometry-based approach (rectangular or circular);
\item
  Using an unsupervised clustering approach;
\item
  Using a custom input, i.e., specifying the blocks with an external
  grouping variable.
\end{itemize}

In some resampling schemes, separation buffers or guard zones can be
imposed to separate the training and test data.

\begin{CodeChunk}
\begin{figure}[ht]

{\centering \includegraphics[width=1\linewidth]{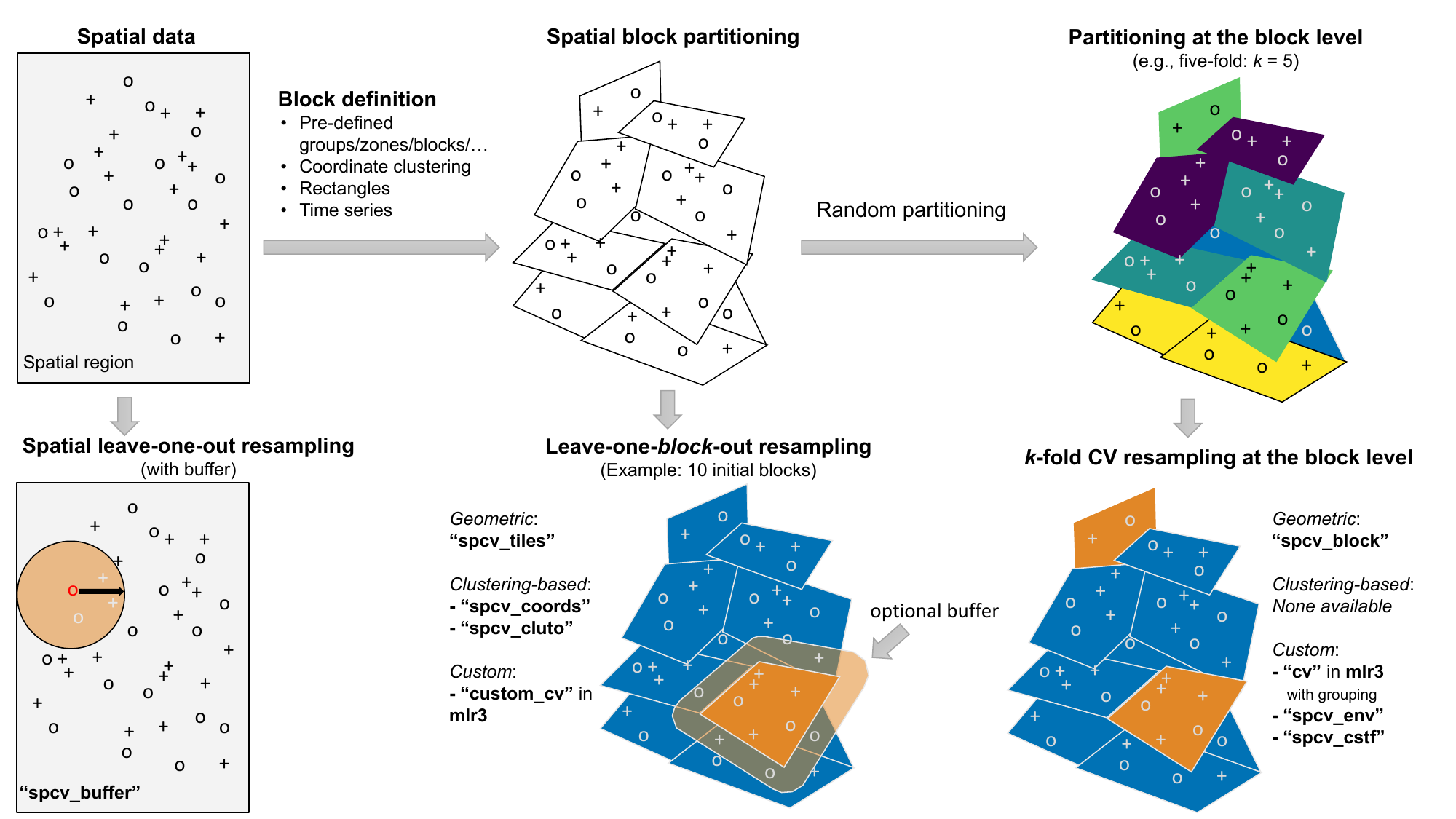}

}

\caption[Conceptual overview of various spatial partitioning schemas]{Conceptual overview of various spatial partitioning schemas. Starting from unpartioned spatial observations (top left) either a 'spatial block partitioning' or a 'spatial leave-one-out resampling' is applied in the first step. A spatial block partitioning can further be turned into a 'leave-one-block-out resampling' or a 'k-fold CV resampling at the block level'. The use of a buffer is theoretically possible in any scenario but in practice only offered by specific method implementations.}\label{fig:rsmp-schema}
\end{figure}
\end{CodeChunk}

\pkg{mlr3spatiotempcv} currently implements the partitioning methods
identified in Table \ref{tab:sptcv-methods}. Several of the implemented
algorithms are themselves versatile toolboxes with multiple options.
Comprehensive and up-to-date information can be found in the package's
online documentation (\url{https://mlr3spatiotempcv.mlr-org.com}). The
following sections give an overview of most implemented partitioning
strategies and their visualization options. The available methods are
further discussed in Section \ref{sec:disc}.

\% this is required to not expand the authors in the table citations on
first use
\shortcites{ploton2020, karasiak2021, moller2021, endicott2017, wu2020, geiss2017, morera2021, zurell2020, brenning2015, bebber2017, jensen2021, escobar2021, stewart2021, reitz2021, egli2020, gao2019}

\begin{table}[ht]
  \centering
  \footnotesize
  \begingroup
    \begin{tabular}{lllll}
      \\
      Type                            & Sub-type & Name                        & \proglang{R} package   & References \\
      \toprule
      \multirow{4}{*}{%
  \begin{tabular}[l]{@{}l@{}}Spatial \\ leave-one-\\out\end{tabular}}        & single point, with buffer &  \texttt{"spcv\_buffer"} & \pkg{blockCV} (2) & %
  \begin{tabular}[l]{@{}l@{}}\cite{ploton2020} \\ \cite{diesing2020}\end{tabular} \\
      \cmidrule{2-5}
       & disc, with buffer  & \texttt{"spcv\_disc"}  & \pkg{sperrorest} (3) & %
  \begin{tabular}[l]{@{}l@{}}\cite{karasiak2021} \\ \cite{moller2021} \\ \cite{endicott2017}\end{tabular} \\
      \midrule
      \multirow{4}{*}{%
  \begin{tabular}[l]{@{}l@{}}Leave-one-\\block-out \\ CV\end{tabular}} &  clustering of coordinates & \texttt{"spcv\_coords"} & \pkg{sperrorest} (6)      & %
  \begin{tabular}[l]{@{}l@{}}\cite{morera2021} \\ \cite{geiss2017} \\ \cite{wu2020}\end{tabular} \\
            \cmidrule{2-5}
                                      &  geometric: rectangular & \texttt{"spcv\_tiles"} & \pkg{sperrorest}    & %
  \begin{tabular}[l]{@{}l@{}}\cite{bebber2017} \\ \cite{zurell2020} \\ \cite{brenning2015}\end{tabular}  \\
                                                  \cmidrule{2-5}
                                      & custom  &  \texttt{"custom\_cv"}           & \pkg{mlr3} (0) & - \\
      \midrule
      \multirow{4}{*}{%
  \begin{tabular}[l]{@{}l@{}}CV at the \\ block level\end{tabular}} &  geometric: rectangular & \texttt{"spcv\_block"}  & \pkg{blockCV} (28)      & %
  \begin{tabular}[l]{@{}l@{}}\cite{jensen2021} \\ \cite{escobar2021} \\ \cite{stewart2021}\end{tabular} \\
  \cmidrule{2-5}
                                      &  custom & \texttt{"cv"} with grouping  & \pkg{mlr3} (0)  &  -  \\
  \cmidrule{2-5}
                                      &  clustering in feature space & \texttt{"spcv\_env"} & \pkg{blockCV} (1)   &  \cite{morera2021}  \\
      \midrule
      \multirow{3}{*}{%
  \begin{tabular}[l]{@{}l@{}}Spatiotemp. \\ CV\end{tabular}} &  custom & \texttt{"sptcv\_cstf"} & \pkg{CAST} (6)      & %
  \begin{tabular}[l]{@{}l@{}}\cite{gao2019} \\ \cite{reitz2021} \\ \cite{egli2020}\end{tabular} \\
                                                  \cmidrule{2-5}
                                      &  clustering: custom & \texttt{"sptcv\_cluto"}  & \pkg{skmeans} (0)    & - \\
                                      \bottomrule
    \end{tabular}
  \endgroup
  \caption[b]{Available spatiotemporal resampling methods in the \pkg{mlr3} ecosystem.
    The "Name" column shows the \pkg{mlr3} method name as found in the \texttt{mlr3::mlr\_resamplings} dictionary.
    The count in brackets after the package name represents the number of studies that were found having used this resampling technique until May 2021.
    For each method, up to three randomly selected references were added to the table.}
    \label{tab:sptcv-methods}
\end{table}

Users are encouraged to contribute new or missing spatiotemporal
resampling methods directly to \pkg{mlr3spatiotempcv}. The already
implemented methods can be inspected to get to know the class structure,
active bindings and methods.

\hypertarget{spatial-leave-one-out}{%
\subsection{Spatial leave-one-out}\label{spatial-leave-one-out}}

Spatial leave-one-out methods use individual observations in space as
test partitions and apply circular buffer or guard zones around around
these test points to enforce a minimum prediction distance.
Leave-one-disc-out resampling modifies this approach to leave out
circular regions centered at observation points.

\subsubsection[Spatial leave-one-out with buffer --- "spcv_buffer"]{Spatial
leave-one-out with buffer ---
\code{"spcv_buffer"}}\label{spatial-leave-one-out-with-buffer}

Leave-one-out CV with buffer and several adaptations for species
distribution modeling \citep{hijmans2020} are implemented in the
\pkg{blockCV} package as the so-called ``buffering'' method and
integrated into \pkg{mlr3spatiotempcv} under the label
\texttt{"spcv\_buffer"}. In species distribution modeling, the response
variable can either be recorded as presence/absence data or as
presence/background information; both options are supported by this
implementation. By default, the dataset contains confirmed presence and
confirmed absence observations, i.e., locations where a species was
observed and not observed, respectively, and therefore spatial LOO CV in
its usual sense can be carried out. Figure \ref{fig:buffer-eval} shows
the first test fold generated with this method for presence/absence data
with a buffer distance of 1000 m.

\begin{CodeChunk}
\begin{CodeInput}
R> library("mlr3")
R> library("mlr3spatiotempcv")
R> task = tsk("ecuador")
R> rsmp_buffer = rsmp("spcv_buffer", theRange = 1000)
R>
R> autoplot(rsmp_buffer, size = 0.8, task = task, fold_id = 1)
\end{CodeInput}
\end{CodeChunk}

\begin{CodeChunk}
\begin{CodeOutput}
<ResamplingSpCVBuffer>: Spatial buffering resampling
* Iterations: 0
* Instantiated: FALSE
* Parameters: theRange=1000
\end{CodeOutput}
\begin{figure}[ht]

{\centering \includegraphics[width=0.4\linewidth]{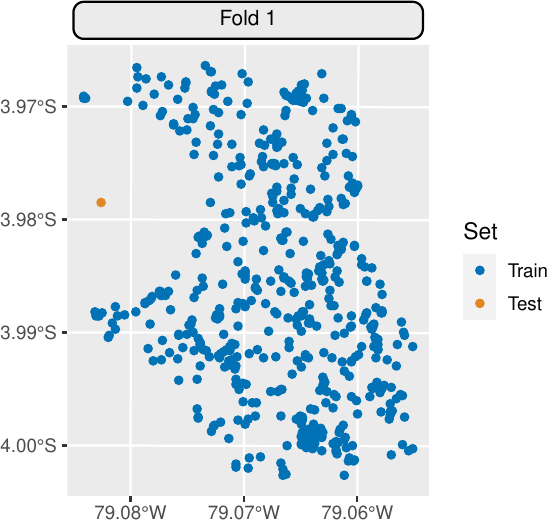}

}

\caption[Visualization of the spatial buffering method from package \pkg{blockCV} (method "spcv\_buffer" in \pkg{mlr3spatiotempcv})]{Visualization of the spatial buffering method from package \pkg{blockCV} (method "spcv\_buffer" in \pkg{mlr3spatiotempcv}). The buffer distance is 1000 m.}\label{fig:buffer-eval}
\end{figure}
\end{CodeChunk}

In the presence/background (or presence-only) situation, in contrast,
only presence observations are recorded, and all other locations within
the study area are referred to as background and considered as
pseudo-absences. Presence/background modeling can be enabled with the
argument \texttt{spDataType\ =\ "PB"}. In this situation, the method
constructs test folds that are centered at the recorded presence
locations, offering two different modes of operation. With
\texttt{addBG\ =\ TRUE} (the default), all background points with a
distance of \texttt{theRange} around a test (presence) point are
included in the test fold as absence data; note that in this case, there
is no separation buffer between training and test samples. The
\texttt{addBG\ =\ FALSE} setting, in contrast, for which no background
data is added to the test fold, then contains only one (presence)
observation, and only the data at a distance of \texttt{theRange} or
greater are included in the training sample, including background data
from these areas.

The application of LOO methods can be computationally expensive since
the method cycles through the entire dataset and fits one model for each
test fold.

\subsubsection[Leave-one-disc-out with optional buffer --- "spcv_disc"]{Leave-one-disc-out
with optional buffer ---
\code{"spcv_disc"}}\label{leave-one-disc-out-with-optional-buffer}

Leave-one-disc-out resampling from package \pkg{sperrorest} defines
circular test sets that are centered at sample locations, and optionally
excludes a buffer zone from the remaining training data. It thus ensures
that a minimum separation distance between training and test data is
maintained. The number of discs is specified by the \texttt{folds}
argument, which defaults to the sample size \(n\). Sample locations are
selected randomly when \texttt{folds} is smaller than \(n\); it is
optionally possible to sample with replacement
(\texttt{replace\ =\ TRUE}). Leave-one-disc-out resampling becomes
spatial LOO CV for a radius of 0 m and when each observation is at a
unique location.

It should be noted that the resampled discs will potentially overlap.
Strictly speaking, this straightforward extension of spatial LOO does
therefore not establish a disjoint partitioning as used for CV
resampling in the traditional sense.

\begin{CodeChunk}
\begin{CodeInput}
R> rsmp_disc = rsmp("spcv_disc", folds = 100, radius = 300L, buffer = 400L)
R> rsmp_disc
R>
R> autoplot(rsmp_disc, size = 0.8, task = task, fold_id = 1)
\end{CodeInput}
\end{CodeChunk}

\begin{CodeChunk}
\begin{CodeOutput}
<ResamplingSpCVDisc>: Repeated Spatial 'disc' resampling
* Iterations: 100
* Instantiated: FALSE
* Parameters: folds=100, radius=300, buffer=400
\end{CodeOutput}
\begin{figure}[ht]

{\centering \includegraphics[width=0.4\linewidth]{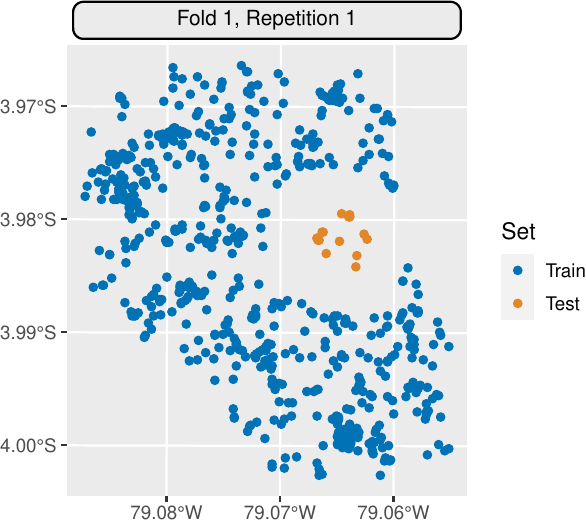}

}

\caption[Visualization of one training set / test set combination generated with the leave-one-disc-out method from package \pkg{sperrorest} (method `"spcv\_disc"` in \pkg{mlr3spatiotempcv})]{Visualization of one training set / test set combination generated with the leave-one-disc-out method from package \pkg{sperrorest} (method `"spcv\_disc"` in \pkg{mlr3spatiotempcv}). The disc has a radius of 300 m and is surrounded by a 400-m buffer.}\label{fig:disc-eval}
\end{figure}
\end{CodeChunk}

\hypertarget{sec:lobo-cv}{%
\subsection{Leave-one-block-out cross-validation}\label{sec:lobo-cv}}

Leave-one-block-out resampling methods partition the dataset spatially
in order to use each of the resulting partitions as a CV test fold.

\subsubsection[Clustering-based: using coordinates --- "spcv_coords"]{Clustering-based:
using coordinates ---
\code{"spcv_coords"}}\label{clustering-based-using-coordinates}

Cluster analysis provides a flexible approach to creating irregularly
shaped spatial blocks for spatial resampling. Numerous techniques are
available that can potentially be applied to the spatial coordinates of
observations, to the features, or to a combination of both. In spatial
model assessment, the focus has been on coordinate-based clustering, and
specifically on leave-one-block-out resampling with blocks created by
\(k\)-means clustering of the coordinates \citep{russ2010}.

Coordinate-based clustering for spatial CV
\citep{russ2010, brenning2012} as implemented in package
\pkg{sperrorest} uses the coordinates of all observations to create
clusters in the spatial domain with the help of the \(k\)-means
clustering algorithm. This can be regarded as a leave-one-block-out
resampling method, or as a \(k\)-fold CV in which each test set is a
spatial cluster. This method is referred to as \texttt{"spcv\_coords"}
in \pkg{mlr3spatiotempcv}.

The coordinate-based clustering approach is very versatile as it adapts
to irregularly-shaped study areas and ensures that exactly \(k\)
partitions are created, which are usually of very similar size when the
sample locations are spread out evenly. Nevertheless, despite the random
selection of initial cluster centers, repeated partitionings may in some
cases be nearly identical. Also, \(k\)-means clustering may be less
suitable for data sets with pre-existing clusters of points and/or with
isolated, distant sample locations. When distinct clusters of points are
present, as in multi-level sampling, it may be better to define clusters
using a factor variable (see method \texttt{"custom\_cv"} in Section
\ref{sec:custom-cv}).

\begin{CodeChunk}
\begin{CodeInput}
R> rsmp_coords = rsmp("spcv_coords", folds = 5)
R>
R> autoplot(rsmp_coords, size = 0.8, fold_id = 1, task = task)
\end{CodeInput}
\end{CodeChunk}

\begin{CodeChunk}
\begin{figure}[ht]

{\centering \includegraphics[width=0.4\linewidth]{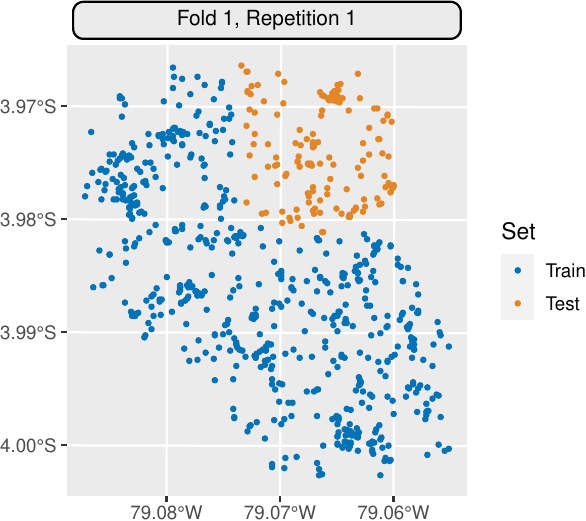}

}

\caption[Leave-one-block-out CV based on $k$-means clustering of the coordinates as implemented in package \pkg{sperrorest} (method `"spcv\_coords"` in \pkg{mlr3spatiotempcv})]{Leave-one-block-out CV based on $k$-means clustering of the coordinates as implemented in package \pkg{sperrorest} (method `"spcv\_coords"` in \pkg{mlr3spatiotempcv}).}\label{fig:coords-eval}
\end{figure}
\end{CodeChunk}

\subsubsection[Geometric: using rectangular blocks --- "spcv_tiles"]{Geometric:
using rectangular blocks ---
\code{"spcv_tiles"}}\label{geometric-using-rectangular-blocks}

Leave-one-tile-out resampling is implemented in the
\texttt{"spcv\_tiles"} method imported from package \pkg{sperrorest}. It
uses rectangular blocks that can be rotated (argument
\texttt{rotation}), and a minimum number or fraction of observations per
block can optionally be achieved by iteratively merging small blocks
into adjacent blocks (argument \texttt{reassign} in conjunction with
\texttt{min\_n} or \texttt{min\_frac}). Block size or number is
specified via the argument \texttt{dsplit} or \texttt{nsplit},
respectively, and square blocks can be obtained with a single (or two
identical) \texttt{dsplit} value(s).

Note that the actual number of folds obtained may be smaller than
\texttt{nsplit{[}1{]}*nsplit{[}2{]}} (or smaller than what would be
expected based on \texttt{dsplit}) since some blocks may be empty or
(optionally) merged into adjacent folds. In the example, there are only
eleven folds instead of twelve because the southwestern part of the
study area's bounding box does not contain observations (Figure
\ref{fig:tile}).

\begin{CodeChunk}
\begin{CodeInput}
R> requireNamespace("sperrorest")
R> rsmp_tiles = rsmp("spcv_tiles", nsplit = c(3L, 4L))
R>
R> autoplot(rsmp_tiles, size = 0.8, fold_id = 1, task = task)
\end{CodeInput}
\end{CodeChunk}

\begin{CodeChunk}
\begin{figure}[ht]

{\centering \includegraphics[width=0.4\linewidth]{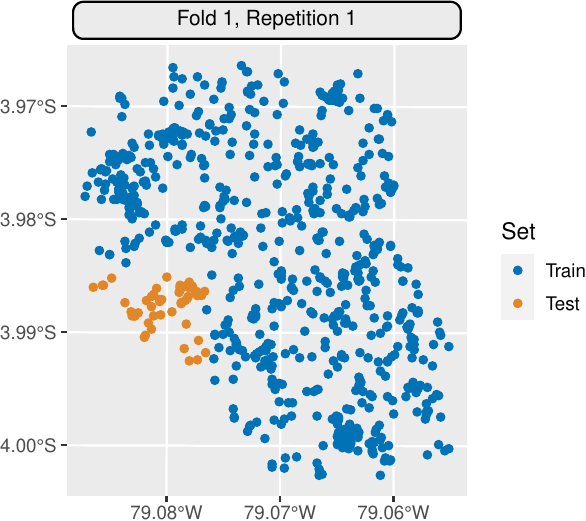}

}

\caption[Leave-one-block-out resampling from package \pkg{sperrorest} (method `"spcv\_tiles"` in package \pkg{mlr3spatiotempcv} with argument `nsplit = c(3,4)` indicating the number of rows and columns)]{Leave-one-block-out resampling from package \pkg{sperrorest} (method `"spcv\_tiles"` in package \pkg{mlr3spatiotempcv} with argument `nsplit = c(3,4)` indicating the number of rows and columns).}\label{fig:tile}
\end{figure}
\end{CodeChunk}

\subsubsection[Custom: "custom_cv" in mlr3]{Custom:
\texttt{"custom\_cv"} in \pkg{mlr3}}\label{sec:custom-cv}

Support for user-defined partitioning strategies is built into
\pkg{mlr3} directly. In this so-called ``Custom CV'', users supply a
factor variable, each level of which defines a partition. The factor
variable can either be specified through a factor vector of the same
length as number of observations, or by passing the name of a feature
within the task (argument \texttt{col}). The following simple example
(taken from \texttt{sperrorest::partition\_factor()}) creates
altitudinal zones that define the spatial partitions.

\begin{CodeChunk}
\begin{CodeInput}
R> breaks = quantile(task$data()$dem, seq(0, 1, length = 6))
R> zclass = cut(task$data()$dem, breaks, include.lowest = TRUE)
R>
R> rsmp_custom = rsmp("custom_cv")
R> rsmp_custom$instantiate(task, f = zclass)
R>
R> autoplot(rsmp_custom, size = 0.8, task = task, fold_id = 1)
\end{CodeInput}
\end{CodeChunk}

\begin{CodeChunk}
\begin{figure}[ht]

{\centering \includegraphics[width=0.4\linewidth]{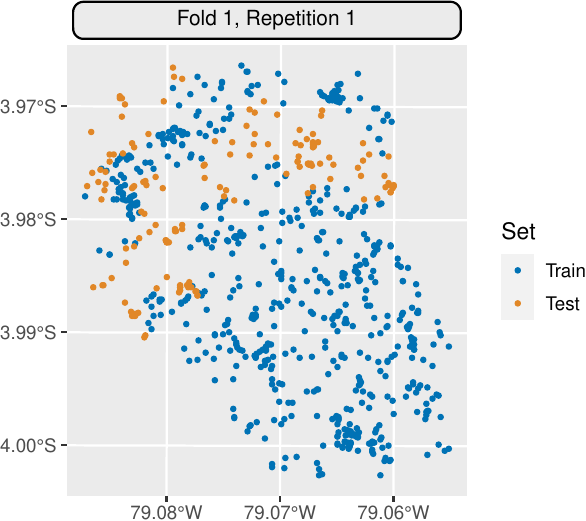}

}

\caption[Leave-one-level-out (custom) resampling from package \pkg{mlr3} (method `"custom\_cv"`)]{Leave-one-level-out (custom) resampling from package \pkg{mlr3} (method `"custom\_cv"`). A factor variable is used to define all partitions.}\label{fig:custom-cv}
\end{figure}
\end{CodeChunk}

\hypertarget{sec:block-cv}{%
\subsection{Cross-validation at the block level}\label{sec:block-cv}}

Methods which operate at the block level first group the observations
into blocks and then combine these blocks into CV partitions. In
\(k\)-fold CV resampling at the block level, there are therefore \(k\)
partitions, each consisting of \(1/k\)-th of the blocks. The special
case in which \(k\) equals the number of blocks, CV at the block level
simply becomes leave-one-block-out CV, for which dedicated
implementations exist (see Section \ref{sec:lobo-cv}).

\subsubsection[Geometric: using rectangular blocks --- "spcv_block"]{Geometric:
using rectangular blocks ---
\code{"spcv_block"}}\label{geometric-using-rectangular-blocks-1}

The \texttt{"spcv\_block"} method from package \pkg{blockCV} supports
both random and systematic resampling of square blocks with argument
\texttt{selection\ =\ "random"} and \texttt{"systematic"}, respectively;
(see Figure \ref{fig:block-random} and Figure
\ref{fig:block-systematic}). There are additional options for modeling
presence-only data, which is a typical use case in species distribution
modeling. Users can furthermore supply a user-defined polygon via
argument \texttt{rasterLayer} with predefined blocking zones.

The size of the square blocks (in meters) are determined by the
\texttt{range} argument. Rectangular blocks can be created by specifying
the number of desired rows and columns (arguments \texttt{rows} and
\texttt{cols}). Due to the non-trivial specification of argument
\texttt{range} package \pkg{blockCV} provides the helper functions
\texttt{spatialAutoRange()} and \texttt{rangeExplorer()} to conduct a
data-driven estimation of the distance at which the spatial
autocorrelation within the data levels off \citep{blockCV}. According to
the package authors, this estimate should then be used for argument
\texttt{range} to have a sensible value for the block sizes created in
method \texttt{"spcv\_block"}.

It should be noted that rectangular partitioning can be problematic in
irregularly shaped study areas as shown in Figure \ref{fig:block-random}
where some of the resulting partitions may contain substantially fewer
observations than others.

\begin{CodeChunk}
\begin{CodeInput}
R> rsmp_block_random = rsmp("spcv_block", range = 1000, folds = 5)
R>
R> autoplot(rsmp_block_random, size = 0.8, fold_id = 1, task = task,
+   show_blocks = TRUE, show_labels = TRUE)
\end{CodeInput}
\end{CodeChunk}

\begin{CodeChunk}
\begin{figure}[ht]

{\centering \includegraphics[width=0.4\linewidth]{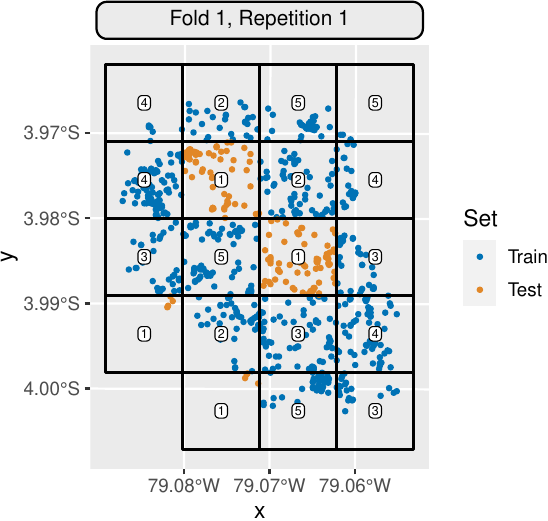}

}

\caption[Random resampling of square spatial blocks using the implementation in package \pkg{blockCV} (method `"spcv\_block"` with option `selection = "random"` in \pkg{mlr3spatiotempcv})]{Random resampling of square spatial blocks using the implementation in package \pkg{blockCV} (method `"spcv\_block"` with option `selection = "random"` in \pkg{mlr3spatiotempcv}). The size of the squares is 1000 m, and four out of the 19 blocks were assigned to the test partition.}\label{fig:block-random}
\end{figure}
\end{CodeChunk}

In systematic resampling, the blocks are numbered row by row, and blocks
\(i+j\cdot\texttt{folds}\) are assigned to fold \(i\) (see Figure
\ref{fig:block-systematic}). This may create undesired patterns when the
number of columns is equal to or a multiple of the number of folds.

\begin{CodeChunk}
\begin{CodeInput}
R> rsmp_block_systematic = rsmp("spcv_block",
+   range = 1000, folds = 5, selection = "systematic"
+ )
R>
R> autoplot(rsmp_block_systematic, size = 0.8, fold_id = 1, task = task,
+   show_blocks = TRUE, show_labels = TRUE)
\end{CodeInput}
\end{CodeChunk}

\begin{CodeChunk}
\begin{figure}[ht]

{\centering \includegraphics[width=0.4\linewidth]{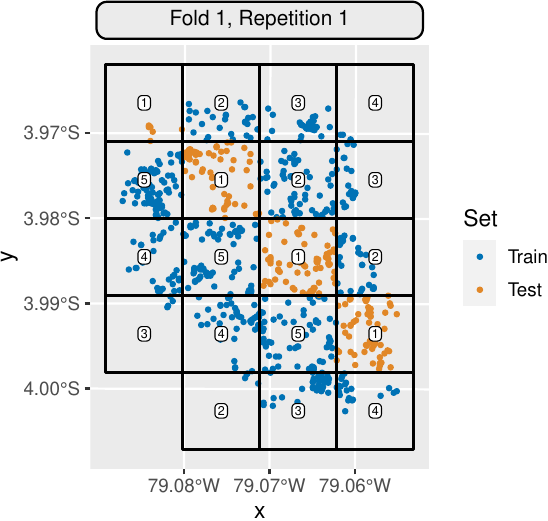}

}

\caption[Sytematic resampling of square spatial blocks using the implementation in package \pkg{blockCV} (method `"spcv\_block"` with option `selection = "systematic"` in \pkg{mlr3spatiotempcv})]{Sytematic resampling of square spatial blocks using the implementation in package \pkg{blockCV} (method `"spcv\_block"` with option `selection = "systematic"` in \pkg{mlr3spatiotempcv}). The size of the squares is 1000 m, and four out of the 19 blocks were assigned to this test sample.}\label{fig:block-systematic}
\end{figure}
\end{CodeChunk}

\emph{Checkerboard} partitioning is a special case of a systematic block
partitioning (\texttt{selection\ =\ "checkerboard"}) which is why we
omitted a practical example for this option. It inherently supports only
two folds, making it less appealing than the more commonly used five- or
ten-fold resampling, which achieve larger training set sizes.

\subsubsection[Custom: "cv" with grouping in mlr3]{Custom: \texttt{"cv"}
with grouping in \pkg{mlr3}}\label{sec:custom-cv-grouping}

Although the \texttt{"cv"} resampling strategy in \pkg{mlr3} performs
random, non-spatial partitioning by default, it can also be used for CV
at the block level. This is achieved by specifying the ``group'' column
role in a \pkg{mlr3} \texttt{Task} object, which uses the factor levels
as blocks. A complete group or block of observations is therefore
assigned to a specific partition, which consequently honors the grouping
structure. In the deprecated \pkg{mlr} package this concept was referred
to as ``blocking''.

In contrast to geometric or clustering-based blocks, the spatial or
temporal location is not used explicitly, but rather implicitly through
the spatial or spatiotemporal footprint of each user-defined block.

The following example uses \(k\)-means clustering to generate classes
that are used as blocks. To underline the honoring of the groups, a
number of groups (eight) that is not a multiple of the number of folds
(three) was chosen. The test sets in the first and second folds are
therefore composed of three groups while the third one holds two groups.

\begin{CodeChunk}
\begin{CodeInput}
R> task_cv = tsk("ecuador")
R> group = as.factor(kmeans(task$coordinates(), 8)$cluster)
R> task_cv$cbind(data.frame("group" = group))
R> task_cv$set_col_roles("group", roles = "group")
R>
R> rsmp_cv_group = rsmp("cv", folds = 3)$instantiate(task_cv)
R>
R> print(rsmp_cv_group$instance)
\end{CodeInput}
\end{CodeChunk}

\begin{CodeChunk}
\begin{CodeInput}
R> autoplot(rsmp_cv_group, size = 0.8, task = task_cv, fold_id = 1)
\end{CodeInput}
\end{CodeChunk}

\begin{CodeChunk}
\begin{CodeOutput}
   row_id fold
1:      8    1
2:      5    1
3:      6    1
4:      1    2
5:      2    2
6:      3    2
7:      4    3
8:      7    3
\end{CodeOutput}
\end{CodeChunk}

\begin{CodeChunk}
\begin{figure}[ht]

{\centering \includegraphics[width=0.4\linewidth]{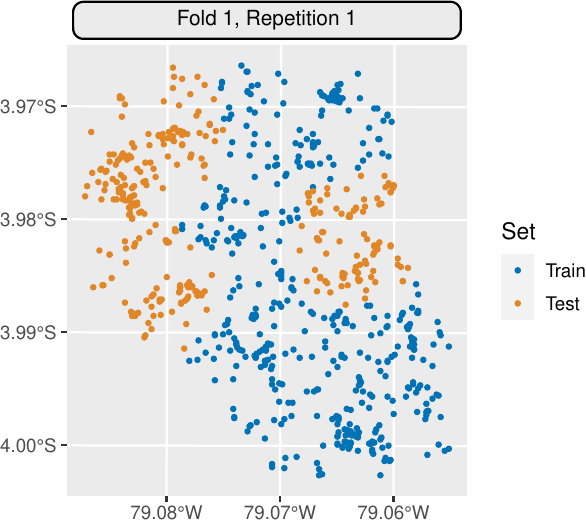}

}

\caption[Cross-Validation at the block level including predefined groups from package \pkg{mlr3} (method `"cv"`)]{Cross-Validation at the block level including predefined groups from package \pkg{mlr3} (method `"cv"`). A factor variable is used to define the grouping. Each class is either assigned to the test or training set.}\label{fig:cv-eval-2}
\end{figure}
\end{CodeChunk}

\subsubsection[Clustering: using feature-based clustering --- "spcv_env"]{Clustering:
using feature-based clustering ---
\code{"spcv_env"}}\label{clustering-using-feature-based-clustering}

The last method from the \pkg{blockCV} package, referred to as
``environmental blocking'' \citep{roberts2017}, makes use of
\emph{k-means} clustering \citep{hartigan1979a} in a possibly
multivariate space to define blocks for resampling at the block level.
The user can select one or multiple numeric features via argument
\texttt{feature} from which the clusters are created. Hereby,
\(k\)-means will use Euclidean distance. To avoid a potential bias
introduced by features with high variance when selecting multiple
features, all features are standardized by default.

In the following example, the observations are clustered based on the
feature ``distance to forest'' (left sub-figure of Figure
\ref{fig:env-1-eval}), which results in a distance-based zonification.
This method also allows to use multiple features for clustering. The
right sub-figure of Figure \ref{fig:env-1-eval} shows the outcome when
using ``distance to deforestation'' and ``slope angle''.

\begin{CodeChunk}
\begin{CodeInput}
R> rsmp_env = rsmp("spcv_env", features = "distdeforest", folds = 5)
R>
R> rsmp_env_multi = rsmp("spcv_env", features = c("distdeforest", "slope"),
+   folds = 5)
R>
R> plot_env_single = autoplot(rsmp_env, size = 0.3,
+   fold_id = 1, task = task) +
+
+   plot_env_multi = autoplot(rsmp_env_multi, size = 0.3,
+     fold_id = 1, task = task)
R>
R> library("patchwork")
R> plot_env_single + plot_env_multi
\end{CodeInput}
\end{CodeChunk}

\begin{CodeChunk}
\begin{figure}[ht]

{\centering \includegraphics[width=1\linewidth]{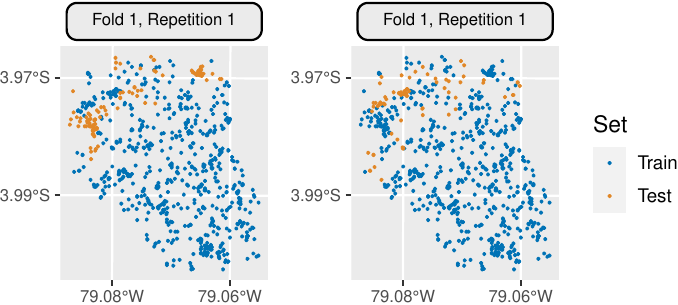}

}

\caption[Environmental leave-one-block-out CV from package \pkg{blockCV} using one (left, `"distdeforest"`) and two (right, `"distdeforest"` and `"slope"`) predictors to define blocks in the feature space]{Environmental leave-one-block-out CV from package \pkg{blockCV} using one (left, `"distdeforest"`) and two (right, `"distdeforest"` and `"slope"`) predictors to define blocks in the feature space. Due to feature space clustering observations are not (necessarily) grouped in the spatial domain.}\label{fig:env-1-eval}
\end{figure}
\end{CodeChunk}

\hypertarget{cross-validation-for-spatiotemporal-data}{%
\subsection{Cross-validation for spatiotemporal
data}\label{cross-validation-for-spatiotemporal-data}}

Some of the implemented resampling methods operate in multiple
dimensions, i.e., in space, time, or space--time. In this section, only
examples of these methods in the spatiotemporal domain will be shown.
For their application in lower dimensions, usually only either the space
or time coordinates need to be omitted from the user input.

\hypertarget{custom-leave-location-and-time-out-and-related-methods}{%
\subsubsection{Custom: ``Leave-location-and-time-out'' and related
methods}\label{custom-leave-location-and-time-out-and-related-methods}}

\citet{meyer2018} proposed a spatiotemporal resampling method in which a
test set is selected and all observations that correspond to the same
location or time point are omitted from the training sample. This method
is referred to as ``leave-location-and-time-out'' (LLTO) in package
\pkg{CAST}. Additional methods that resample in the temporal and spatial
domain only are named ``leave-time-out'' (LTO) and
``leave-location-out'' (LLO), respectively. Note that despite their
names, LLTO, LTO and LLO are conceptually not leave-\emph{one}-out
methods as they place a certain fraction of observations in the test
set, as in ordinary CV. Also, LTO and LLO are conceptually similar to
\pkg{mlr3}'s ``cv'' method with a custom grouping as they perform a CV
at the block level using a grouping structure defined by time points
(LTO) and locations (i.e., time series; LLO).

In this section the \texttt{cookfarm} dataset is used as an example
because it has a temporal dimension identified by the variable ``Date''.

\texttt{mlr3spatiotempcv::autoplot()} supports two visualization types
for spatiotemporal methods which can be selected via the logical
argument \texttt{plot3D}. The heavy lifting of the 3D visualization
(i.e., 2D + time) option is done via package \pkg{plotly}. Because a
dynamic image cannot be included in this manuscript, static versions,
which can be generated by setting \texttt{static\_image\ =\ TRUE}, are
shown (see for example Figure \ref{fig:lto}).

\paragraph[CV at the time-point level: "leave-time-out" (LTO) --- "sptcv_cstf"]{CV
at the time-point level: ``leave-time-out'' (LTO) ---
\code{"sptcv_cstf"}}\label{cv-at-the-time-point-level-leave-time-out-lto}

In the LTO method, the time points are resampled into the desired number
of folds. In the terminology used in this work, this can be referred to
as resampling at the level of time points, which effectively define
blocks. Thus, observations from the same time point are jointly sampled
into the same test (or training) fold, with no constraints on the
temporal distance between the sampled time points. This method does
therefore not implement block CV in the sense of the time series
literature.

In the \texttt{cookfarm\_mlr3} example dataset, the \texttt{Date}
variable was reduced to five unique levels for better visualization, and
then used to create a spatiotemporal regression task in
\pkg{mlr3spatiotempcv} (Figure \ref{fig:lto}). In \texttt{autoplot()}, a
stratified sample based on the partitions is taken to reduce the number
of points plotted.

\begin{CodeChunk}
\begin{CodeInput}
R> data = cookfarm_mlr3
R> set.seed(42)
R> data$Date = sample(rep(c(
+   "2020-01-01", "2020-02-01", "2020-03-01", "2020-04-01",
+   "2020-05-01"), times = 1, each = 35768))
R> task_spt = as_task_regr_st(data,
+   id = "cookfarm", target = "PHIHOX",
+   coordinate_names = c("x", "y"), coords_as_features = FALSE,
+   crs = 26911)
R> task_spt$set_col_roles("Date", roles = "time")
R>
R> rsmp_cstf_time = rsmp("sptcv_cstf", folds = 5)
R>
R> p_lto = autoplot(rsmp_cstf_time,
+   fold_id = 5, task = task_spt, plot3D = TRUE,
+   point_size = 6, axis_label_fontsize = 15,
+   sample_fold_n = 3000L
+ )
R>
R> p_lto_print = plotly::layout(p_lto,
+   scene = list(camera = list(eye = list(z = 0.58))),
+   showlegend = FALSE, title = "",
+   margin = list(l = 0, b = 0, r = 0, t = 0))
R>
R> plotly::save_image(p_lto_print, "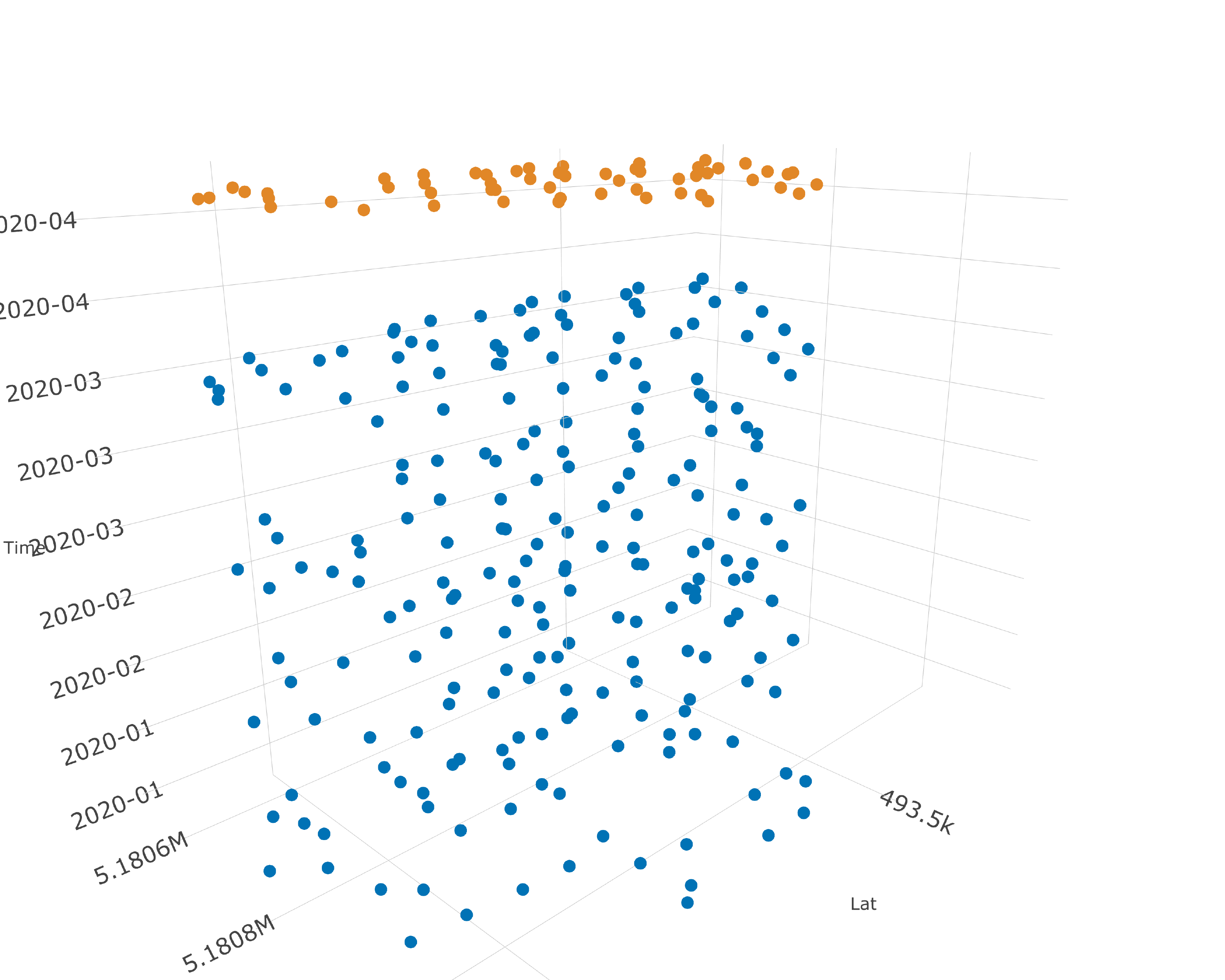",
+   scale = 2, width = 1000, height = 800)
\end{CodeInput}
\end{CodeChunk}

\begin{CodeChunk}
\begin{figure}[ht]

{\centering \includegraphics[width=0.8\linewidth]{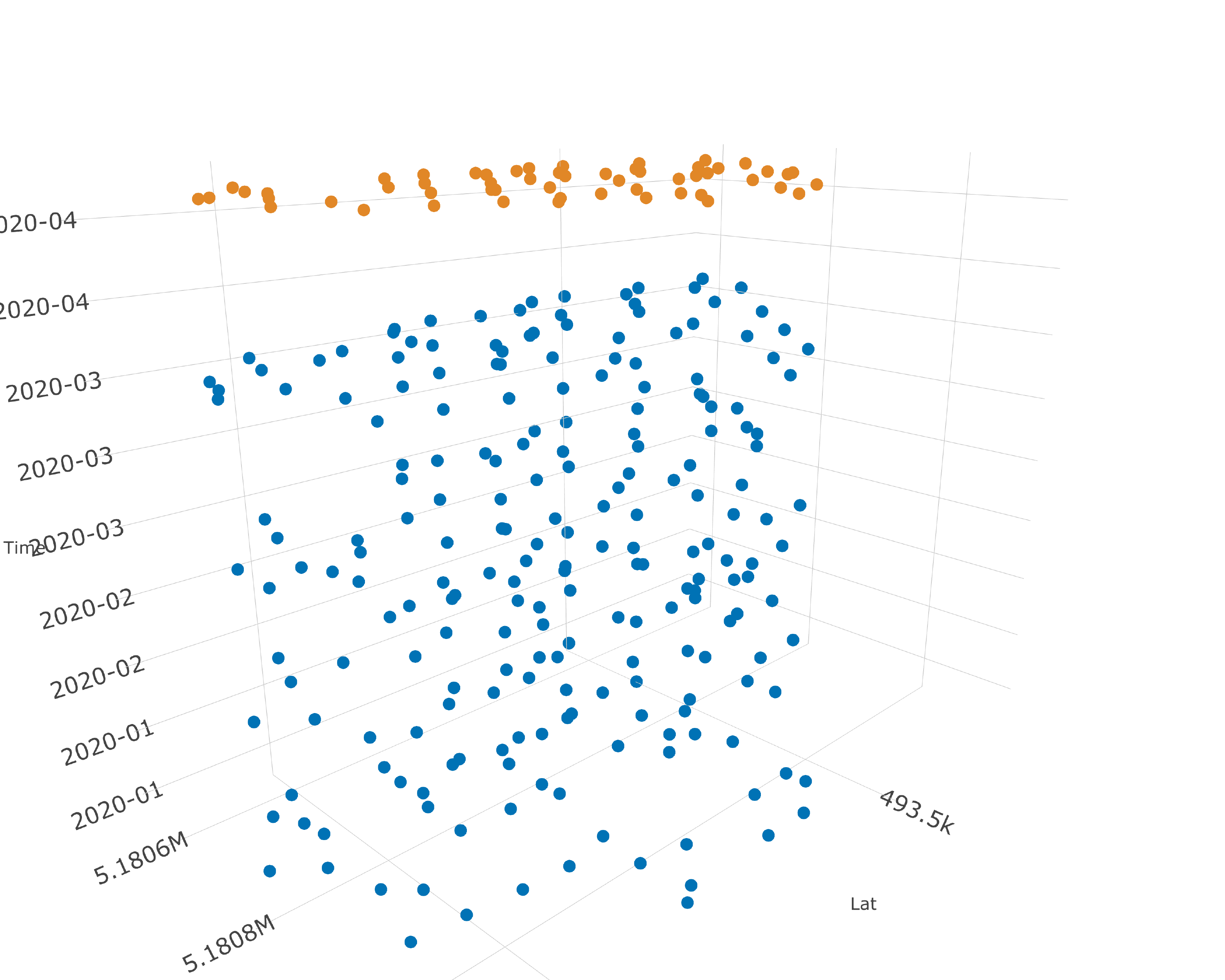}

}

\caption[Perspective plot of `"leave-time-out"` CV from package \pkg{CAST} (method `"sptcv\_cstf"` and column role `"time" = "Date"`)]{Perspective plot of `"leave-time-out"` CV from package \pkg{CAST} (method `"sptcv\_cstf"` and column role `"time" = "Date"`). Only five folds and five time points were used in this example. Note that the blue dots correspond to five discrete time levels, which appear as a point cloud due to the viewing angle.}\label{fig:lto}
\end{figure}
\end{CodeChunk}

\paragraph[CV at the location level: "leave-location-out" (LLO) --- "sptcv_cstf"]{CV
at the location level: ``leave-location-out'' (LLO) ---
\code{"sptcv_cstf"}}\label{cv-at-the-location-level-leave-location-out-llo}

In contrast to LTO, the LLO method randomly resamples locations that
may, for example, correspond to time series. The sampled locations form
the test partition while the temporal information is ignored (Figure
\ref{fig:llo}). Unlike spatial CV methods that are based on geometric
regions or the clustering of coordinates, the sampled test locations
include no particular spatial relationship.

To tell the resampling method to use the `space' column for
partitioning, the `time' column needs to be unset and the `space' column
defined. Because the temporal variable ``Date'' is not in use in this
scenario, \texttt{autoplot()} needs to be instructed explicitily to use
it for 3D plotting via argument \texttt{plot\_time\_var}.

\begin{CodeChunk}
\begin{CodeInput}
R> task_spt$col_roles$time = character()
R> task_spt$set_col_roles("SOURCEID", roles = "space")
R>
R> rsmp_cstf_loc = rsmp("sptcv_cstf", folds = 5)
R>
R> p_llo = autoplot(rsmp_cstf_loc,
+   fold_id = 5, task = task_spt,
+   point_size = 6, axis_label_fontsize = 15,
+   plot3D = TRUE, plot_time_var = "Date",
+   sample_fold_n = 3000L)
R>
R> p_llo_print =
+   plotly::layout(p_llo,
+     scene = list(camera = list(eye = list(z = 2.5, x = -0.1, y = -0.1))),
+     showlegend = FALSE, title = "", polar = TRUE,
+     margin = list(l = 0, b = 0, r = 0, t = 0))
R>
R> plotly::save_image(p_llo_print, "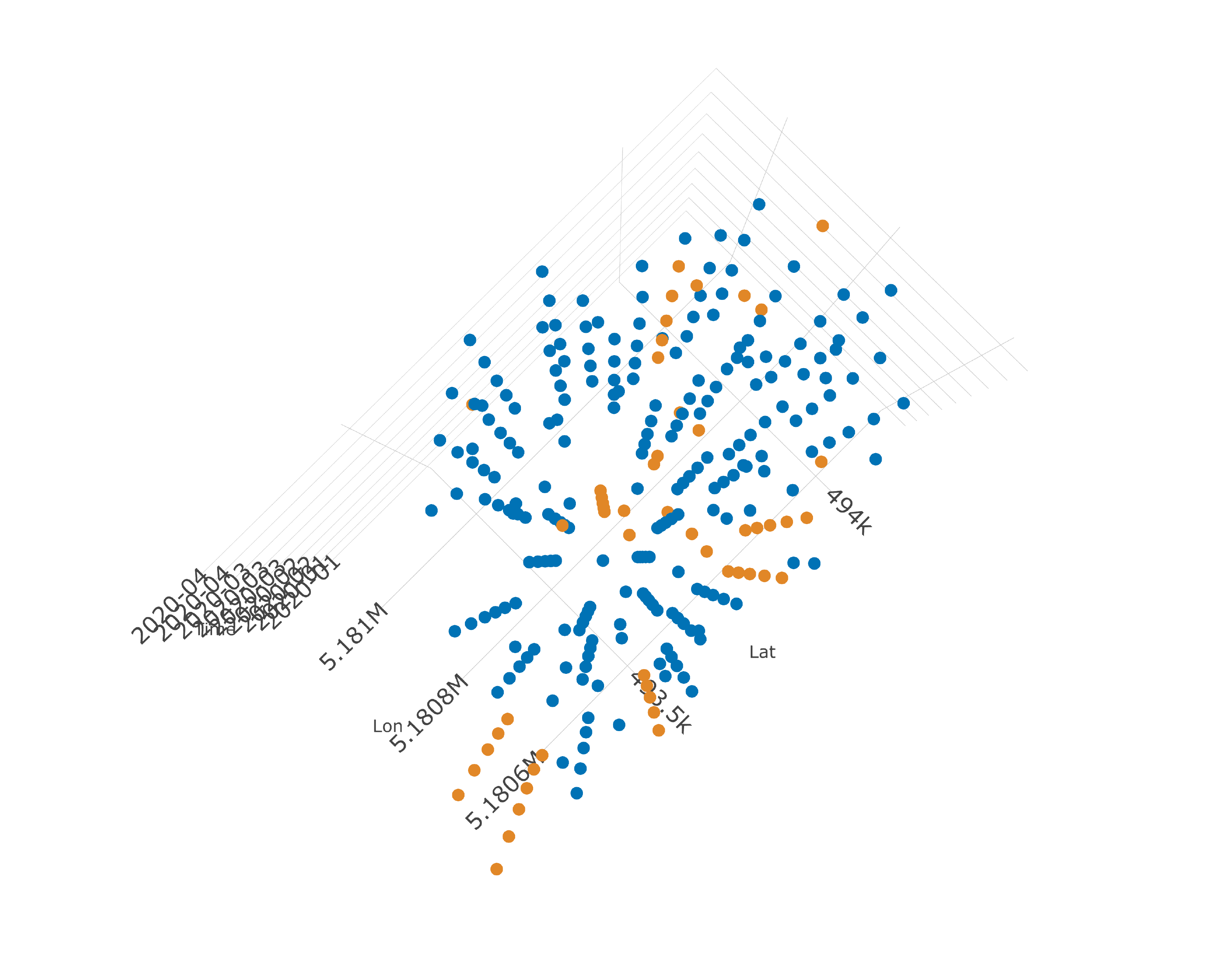",
+   scale = 2, width = 1000, height = 800)
\end{CodeInput}
\end{CodeChunk}

\begin{CodeChunk}
\begin{figure}[ht]

{\centering \includegraphics[width=0.7\linewidth]{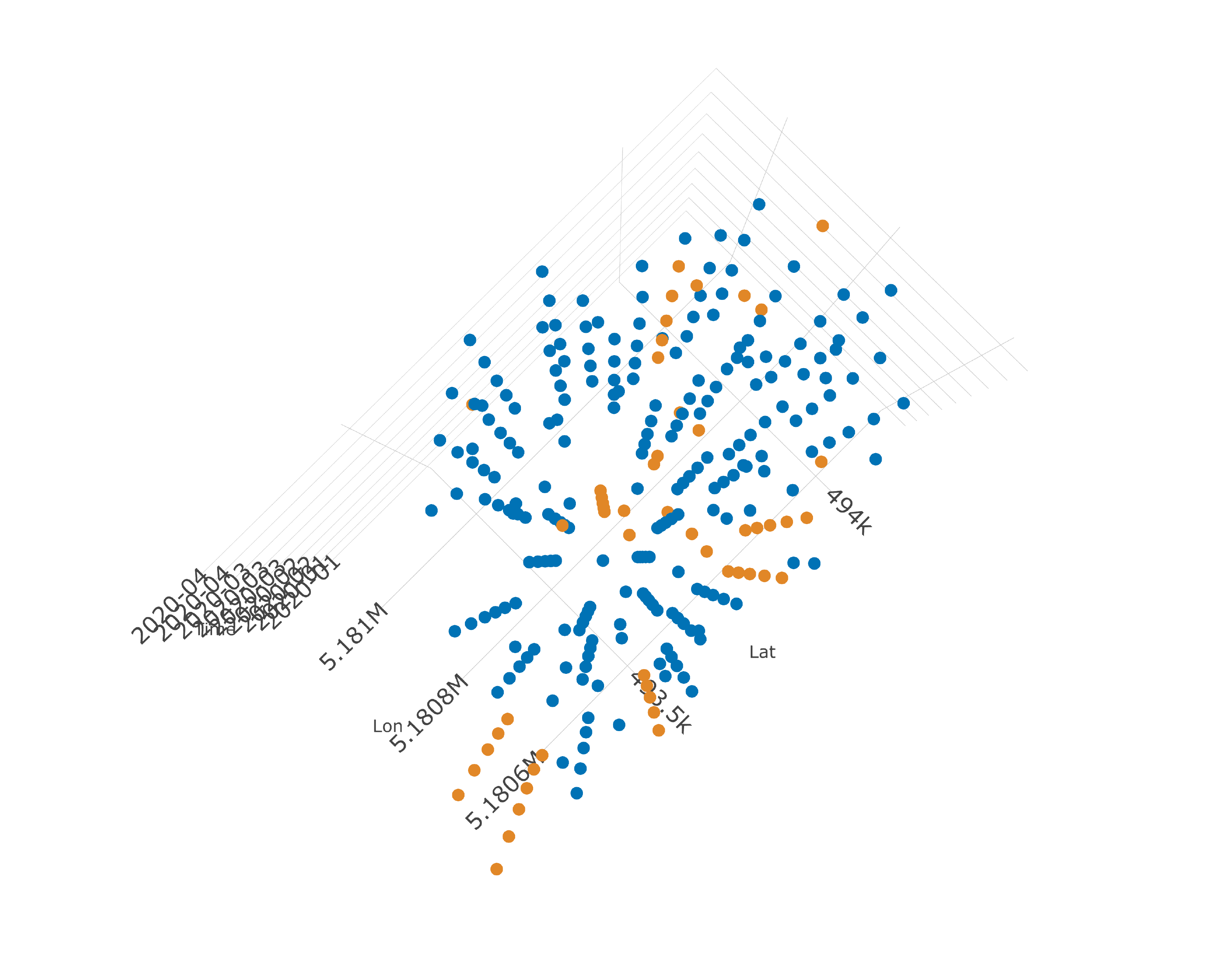}

}

\caption[Birds-eye view of `"leave-location-out"` CV from package \pkg{CAST} (method `"sptcv\_cstf"` and column role `"space" = "SOURCEID"`)]{Birds-eye view of `"leave-location-out"` CV from package \pkg{CAST} (method `"sptcv\_cstf"` and column role `"space" = "SOURCEID"`).}\label{fig:llo}
\end{figure}
\end{CodeChunk}

\paragraph["Leave-location-and-time-out" (LLTO) --- "sptcv_cstf"]{``Leave-location-and-time-out''
(LLTO) --- \code{"sptcv_cstf"}}\label{leave-location-and-time-out-llto}

In LLTO, a test set is first randomly sampled from the data set, and
then all observations that correspond to the same location or time point
are omitted from the training sample (Figure \ref{fig:llto}). LLTO
resampling mimics the situation where a model is trained on time series
data from a number of locations and time points, and used to predict the
time series at other locations and time points that are not included in
the training sample.

Conceptually, LLTO applies zero-distance buffering in both space and
time: The buffer zones consist of all observations whose distance to the
test sample in either space or time equals zero. In a mathematical
sense, however, this buffering is not based on a valid metric (or
distance function) in three-dimensional space (2D + time) as neither the
identity of detectability nor the triangle inequality are satisfied by
the underlying combined `distance' measure. Also note that LLTO does not
`combine' LTO with LLO, as neither of these applies a buffer zone.

The \texttt{"spcv\_cstf"} methods LLO and LTO (with only one of
\texttt{space\_var} or \texttt{time\_var} set) require a variable in the
dataset which should be used for grouping. The specification of the
variable(s) which should be used for a spatial, temporal or
spatiotemporal grouping is not trivial because the final partitioning
should, in the optimal case, ensure that the selected groups inherit
substantial autocorrelation within themselves and simultaneously differ
substantially from other partitions. Also, if the selected variable
contains too many groups, the difference within train/test splits may
become undesirably high and tend towards a LOO CV \citep{meyer2018}.

\begin{CodeChunk}
\begin{CodeInput}
R> task_spt$set_col_roles("SOURCEID", roles = "space")
R> task_spt$set_col_roles("Date", roles = "time")
R>
R> rsmp_cstf_time_loc = rsmp("sptcv_cstf", folds = 5)
R>
R> p_lto = autoplot(rsmp_cstf_time_loc, point_size = 6,
+   axis_label_fontsize = 15,
+   fold_id = 4, task = task_spt, plot3D = TRUE,
+   show_omitted = TRUE, sample_fold_n = 3000L)
R>
R> p_lto_print = plotly::layout(p_lto,
+   scene = list(camera = list(eye = list(z = 0.58))),
+   showlegend = FALSE, title = "",
+   margin = list(l = 0, b = 0, r = 0, t = 0))
R>
R> plotly::save_image(p_lto_print, "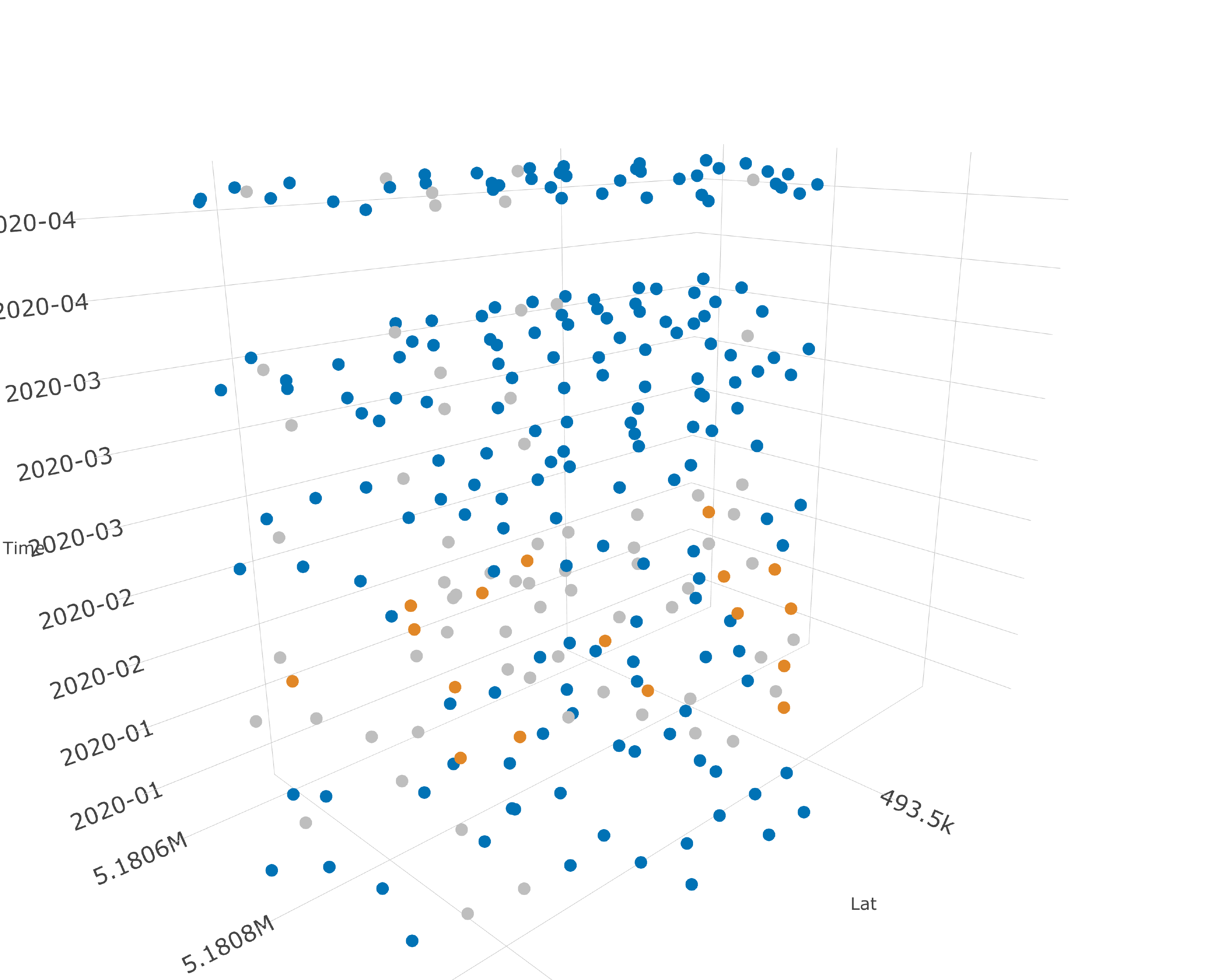",
+   scale = 2, width = 1000, height = 800)
\end{CodeInput}
\end{CodeChunk}

\begin{CodeChunk}
\begin{figure}[h]

{\centering \includegraphics[width=0.7\linewidth]{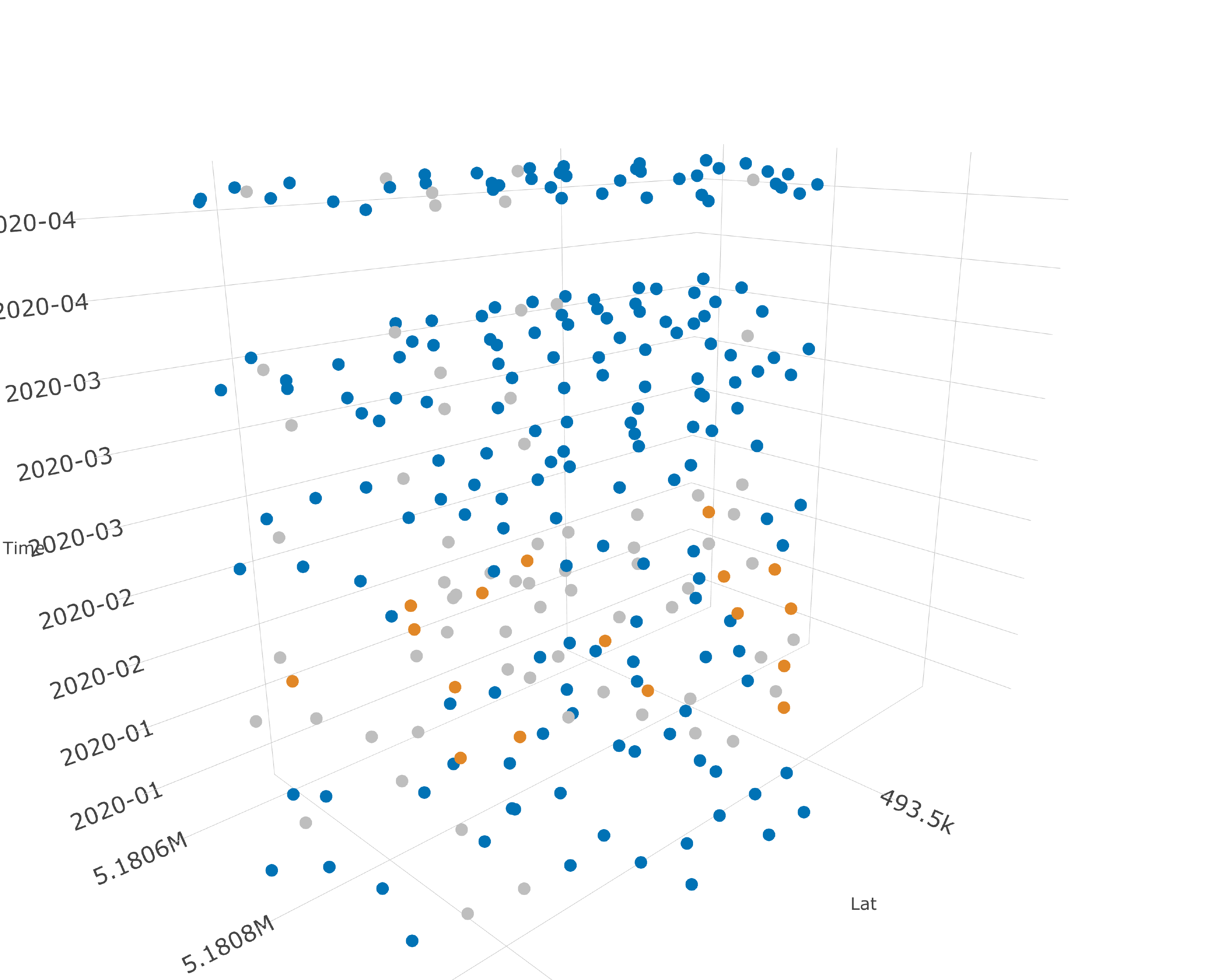}

}

\caption[Perspective plot of `"leave-location-and-time-out"` CV from package \pkg{CAST} (method `"sptcv\_cstf"` and column roles `"time" = "Date"` and `"space" = "SOURCEID"`)]{Perspective plot of `"leave-location-and-time-out"` CV from package \pkg{CAST} (method `"sptcv\_cstf"` and column roles `"time" = "Date"` and `"space" = "SOURCEID"`). The grey points are excluded from both the training and the test set in this example.}\label{fig:llto}
\end{figure}
\end{CodeChunk}

\hypertarget{clustering-using-cluto}{%
\subsubsection{Clustering: using CLUTO}\label{clustering-using-cluto}}

At present, \pkg{mlr3spatiotempcv} also supports spatiotemporal
partitioning using the versatile CLUTO clustering algorithm
\citep{cluto}. CLUTO is available in \proglang{R} through the
\pkg{skmeans} package \citep{skmeans}, which provides an interface to a
downloadable compiled library with a restriction to non-commercial uses
(see
\texttt{help("ResamplingSptCVCluto",\ package\ =\ "mlr3spatiotempcv")}
for more information). Due to this restriction and the age of the latest
release (14 years at the time of writing) this method is not explained
in greater detail.

\hypertarget{sec:case-study}{%
\section{Step-by-step example: Comparing spatial and non-spatial
CV}\label{sec:case-study}}

A well-known case study is used to demonstrate the application of
spatial and non-spatial resampling techniques for model assessment in
\pkg{mlr3spatiotempcv}. The objective of landslide susceptibility
modeling is to predict how prone to landslide initiation a location is.
Models are fitted to historical landslide occurrences, but they need to
learn generalizable relationships between predisposing variables and the
response as opposed to perfectly reproducing or memorizing the
historical distribution. This binary classification task on landslides
in Ecuador \citep{muenchow2012} is available as a built-in task via
\texttt{tsk("ecuador")}, but is generated from the learning sample in
this example. Random forest is used as a classifier, and the area under
the ROC curve (AUROC) as the performance measure.

Spatial CV is implemented in the form of leave-one-block-out CV using
coordinate-based \(k\)-means clustering to generate irregularly shaped
blocks of roughly equal size. This approach is better suited for the
irregular shape of the present study area than a rectangular
partitioning. Figure \ref{fig:vis-spcv-eval} and Figure
\ref{fig:vis-nspcv-eval} show the contrasting distributions of training
and test samples. For demonstration purposes only four CV folds and two
repetitions are used.

Besides the practical example shown below, additional tutorials covering
\pkg{mlr3} use cases can be found at
\href{https://mlr-org.com/gallery}{mlr3gallery}.

\hypertarget{task-preparation}{%
\subsection{Task preparation}\label{task-preparation}}

In \pkg{mlr3}, machine-learning tasks with their respective dataset and
response variable are represented by objects of class \texttt{Task}.
\pkg{mlr3spatiotempcv}'s spatial and spatiotemporal machine-learning
tasks are also derived from this superclass. Specifically, the
\texttt{TaskClassifST} and \texttt{TaskRegrST} classes for
classification and regression tasks require several additional arguments
that must be passed as a named list using the \texttt{extra\_args}
argument:

\begin{itemize}
\tightlist
\item
  \texttt{coordinate\_names}: Names of the features that represent the
  spatial coordinates. This is automatically inferred when a \texttt{sf}
  object is passed.
\item
  \texttt{coords\_as\_features}: Whether the coordinates should be used
  as features; by default they are not.
\item
  \texttt{crs}: The coordinate reference system of the data as a PROJ
  string or EPSG code in the format
  \texttt{ESPG:\textless{}code\textgreater{}}.
\end{itemize}

At first all necessary R packages are loaded and a lower verbosity is
set to keep the output tidy. A random-number seed is set for
reproducibility.

\begin{CodeChunk}
\begin{CodeInput}
R> library("mlr3")
R> library("mlr3spatiotempcv")
R>
R> lgr::get_logger("bbotk")$set_threshold("warn")
R> lgr::get_logger("mlr3")$set_threshold("warn")
R>
R> set.seed(42)
\end{CodeInput}
\end{CodeChunk}

The task \texttt{"ecuador"} is available as an example task in
\pkg{mlr3spatiotempcv} through \texttt{tsk("ecuador")}. To create it
manually from a \texttt{data.frame} named \texttt{ecuador}, one would
do:

\begin{CodeChunk}
\begin{CodeInput}
R> data("ecuador", package = "mlr3spatiotempcv")
R> task = as_task_classif_st(ecuador, target = "slides", positive = "TRUE",
+   coordinate_names = c("x", "y"), coords_as_features = FALSE,
+   crs = "EPSG:32717")
\end{CodeInput}
\end{CodeChunk}

\hypertarget{model-preparation}{%
\subsection{Model preparation}\label{model-preparation}}

Next, the random forest learner (\texttt{"classif.ranger"}) is
initialized with default hyperparameters and the prediction type is set
to \texttt{"probability"} because the model is used for soft
classification. A set of commonly used learners is available in package
\pkg{mlr3learners} \citep{mlr3learners}, including the random forest
implementation of \citet{ranger}.

\begin{CodeChunk}
\begin{CodeInput}
R> library("mlr3learners")
R>
R> learner = lrn("classif.ranger", predict_type = "prob")
\end{CodeInput}
\end{CodeChunk}

\hypertarget{non-spatial-cross-validation}{%
\subsection{Non-spatial
cross-validation}\label{non-spatial-cross-validation}}

To define a resampling strategy, the \texttt{rsmp()} function is used to
generate a resampling object using four folds and two repetitions
following a random sampling logic (``cv'').

Next, the created resampling object \texttt{rsmp\_nsp} is passed to the
\texttt{resample()} function together with the task and learner objects
created earlier to execute the model assessment. This is the actual,
potentially time-consuming CV estimation. With the present settings,
eight random forest classifiers are fitted and evaluated in this step
--- one model fitted on each CV training set.

Model performances are calculated from the CV predictions using the
AUROC (\texttt{"classif.auc"} in \pkg{mlr3} notation).

\begin{CodeChunk}
\begin{CodeInput}
R> rsmp_nsp = rsmp("repeated_cv", folds = 4, repeats = 2)
R> rsmp_nsp
R> rr_nsp = resample(
+   task = task, learner = learner,
+   resampling = rsmp_nsp
+ )
\end{CodeInput}
\end{CodeChunk}

\begin{CodeChunk}
\begin{CodeInput}
R> rr_nsp$aggregate(measures = msr("classif.auc"))
\end{CodeInput}
\begin{CodeOutput}
classif.auc
  0.7600664
\end{CodeOutput}
\end{CodeChunk}

\hypertarget{spatial-cross-validation-via-coordinate-based-clustering}{%
\subsection{Spatial cross-validation via coordinate-based
clustering}\label{spatial-cross-validation-via-coordinate-based-clustering}}

The model assessment is now repeated again using spatial CV resampling,
for which the only required change is to replace \texttt{"repeated\_cv"}
by \texttt{"repeated\_spcv\_coords"}.

\begin{CodeChunk}
\begin{CodeInput}
R> rsmp_sp = rsmp("repeated_spcv_coords", folds = 4, repeats = 2)
R> rsmp_sp
R> rr_sp = resample(
+   task = task, learner = learner,
+   resampling = rsmp_sp
+ )
\end{CodeInput}
\end{CodeChunk}

\begin{CodeChunk}
\begin{CodeInput}
R> rr_sp$aggregate(measures = msr("classif.auc"))
\end{CodeInput}
\begin{CodeOutput}
classif.auc
  0.6100402
\end{CodeOutput}
\end{CodeChunk}

\hypertarget{visualization-of-cv-partitions}{%
\subsection{Visualization of CV
partitions}\label{visualization-of-cv-partitions}}

Finally, we visualize (two of) the partitions that were used during
performance estimation by making use of the generic \texttt{autoplot()}
function in package \pkg{mlr3spatiotempcv} (Figure
\ref{fig:vis-spcv-eval}).

\begin{CodeChunk}
\begin{CodeInput}
R> autoplot(rsmp_sp, task, fold_id = c(1:2), size = 0.8)
\end{CodeInput}
\end{CodeChunk}

\begin{CodeChunk}
\begin{figure}[ht]

{\centering \includegraphics{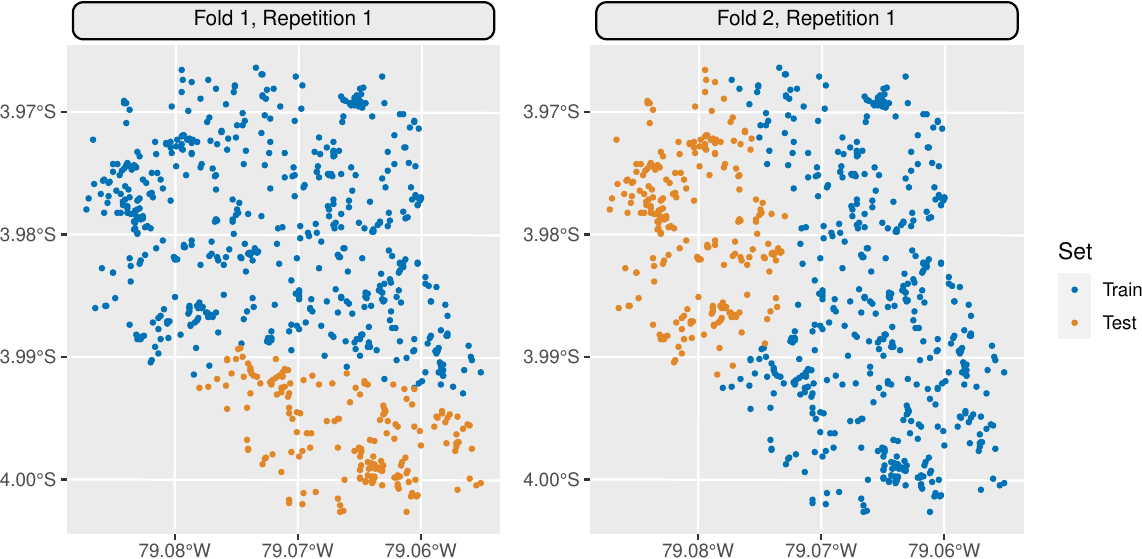}

}

\caption[Spatial leave-one-block-out partitioning using coordinate-based clustering to create roughly equally sized polygonal blocks]{Spatial leave-one-block-out partitioning using coordinate-based clustering to create roughly equally sized polygonal blocks. Due to space limitations only the first two folds of the first repetition are shown.}\label{fig:vis-spcv-eval}
\end{figure}
\end{CodeChunk}

\begin{CodeChunk}
\begin{CodeInput}
R> autoplot(rsmp_nsp, task, fold_id = c(1:2), size = 0.8)
\end{CodeInput}
\end{CodeChunk}

\begin{CodeChunk}
\begin{figure}[ht]

{\centering \includegraphics{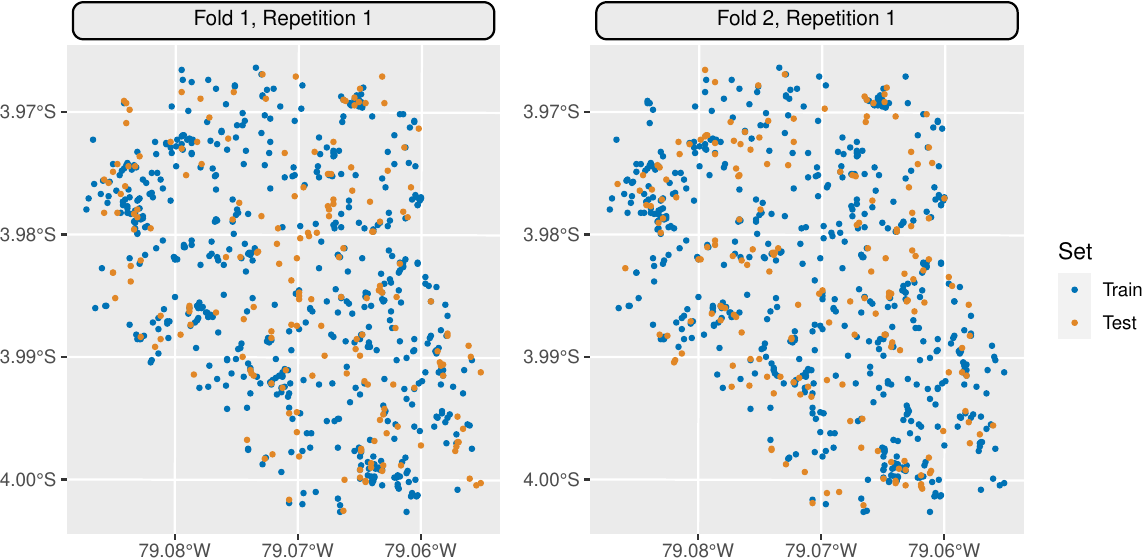}

}

\caption[Random (non-spatial) four-fold CV partitioning]{Random (non-spatial) four-fold CV partitioning. Only the first two folds of the first repetition are shown.}\label{fig:vis-nspcv-eval}
\end{figure}
\end{CodeChunk}

\hypertarget{interpretation}{%
\subsection{Interpretation}\label{interpretation}}

If one takes a closer look at the results, the non-spatial CV estimate
of AUC (0.76) is substantially higher compared to the spatial CV
estimate of 0.64. Since test points in non-spatial CV may be from the
same slopes or even the same landslides as the training data, the
non-spatial CV result should/can be considered as an over-optimistic
estimate of the model's ability to predict the susceptibility to ``new''
landslides. Spatial CV, in contrast, provides a better/more accurate
measure of a model's ability to generalize from the training sample ---
in this case study, from the specific hillslopes and historical
landslides in the training sample. It is also expected that spatial CV
results better represent the model's transferability to geologically and
topographically similar areas adjacent to the training area. The
magnitude of the difference between spatial and non-spatial CV estimates
may depend on the dataset, the strength on spatial or spatiotemporal
autocorrelation, and the learner itself. Algorithms with a higher
tendency to overfit to the training set will tend to have a larger
spread in such scenarios.

\hypertarget{sec:disc}{%
\section{Discussion}\label{sec:disc}}

\hypertarget{choosing-a-resampling-method-for-model-assessment}{%
\subsection{Choosing a resampling method for model
assessment}\label{choosing-a-resampling-method-for-model-assessment}}

The question of which resampling method should be chosen for a
prediction task and dataset at hand comes up regularly in practice. Even
though there is and most likely will be no definitive answer to this
question, we would like to give some guidance in this section to help
find an appropriate method. As a general rule, we recommend to use a
resampling scheme that (1) mimics the predictive situation in which the
model will be applied operationally, and (2) is consistent with the
structure of the data. Both aspects are outlined in this section,
starting with two concrete modeling scenarios.

Although the case study example in Section \ref{sec:case-study} used the
\texttt{"spcv\_coords"} method for coordinate-based clustering, this
should not give the impression that this method is the only method
suitable for this example task. In this application setting, we want to
assess how well the model generalized from the concrete set of
historical landslide occurrences, which is why we ensured that training
and test sets contain different, ``new'' hillslopes and landslides.
Coordinate-based clustering is particularly appealing in this setting
because of its ability to adapt to the irregularly shaped study area of
this example. Resampling at the level of sub-catchments could have been
a viable alternative approach that can be implemented using custom
resampling (\texttt{"custom\_cv"} method); however, this may result in
less balanced sizes of test sets as catchment sizes may vary. When the
timing of landslides is known (event-based inventories) or multiple
inventories have been compiled for different time points, it can also be
recommendable to additionally sample training and test data from
different time points, as with the LLTO and LTO \citep{meyer2018} or
similar methods \citep{brenning2005}.

In other application scenarios such as the crop classification example
given in Section \ref{sec:intro} \citep{pena2015}, the objective is to
predict the fruit-tree crop type of `new', unseen fields within an
agricultural region. In this use case we are not at all interested in
``predicting'' the already known crop type of (other) grid cells within
the same agricultural field. Also, a model will likely be much better
able to predict the crop type within the same field from multitemporal
remote-sensing data since all crops within that field are subject to
identical management practices (e.g., use of pesticides, pruning of
fruit trees, tree spacing), while `new' fields may be managed
differently. As a consequence, grid cells from the same field should be
grouped into a block, and resampling should be done at the field level
(\texttt{"cv"} method with grouping) to receive an honest estimate of
the model's performance in a relevant predictive situation. If, in
contrast, the objective is to apply the model to an adjacent
agricultural region (e.g., adjacent county) where the same crop types
are present, it may be advisable to use coordinate-based clustering
(\texttt{"spcv\_coords"} method) to obtain larger, contiguous test
regions.

In summary, there are various factors that may be considered in judging
the suitability of a resampling method:

\begin{itemize}
\tightlist
\item
  Will the model be applied to predict `new' outcomes at near or more
  distant spatial locations?
\item
  Will it be applied to predict into the future, or hindcast gaps
  between spatiotemporal observations in the past?
\item
  Is it necessary to impose a separation distance or prediction horizon
  as a spatial or temporal buffer between training and prediction
  locations?
\item
  How densely are the observations distributed in space and time? Are
  they more densely distributed than the intended spatial or temporal
  prediction distance?
\item
  Is the data naturally grouped, e.g., because of the spatial extent of
  the studied objects, or as a consequence of multi-level (cluster)
  sampling?
\item
  With an eye on environmental blocking and extrapolation in feature
  space, is it intended to apply the model to predict `new' outcomes for
  unobserved values of predictor variables?
\end{itemize}

Based on these criteria users may choose a matching resampling method
that is either more restrictive (by discarding nearby observations for
fold creation) or more liberal (by not removing observations and
eventually ignoring natural grouping patterns). The specific
publications related to the methods integrated into
\pkg{mlr3spatiotempcv} may give further advice and provide additional
use cases for the application of each respective approach. Users should
therefore also refer to publications that are referenced of linked in
the help files of this package or its respective upstream packages.

\hypertarget{resampling-for-hyperparameter-tuning}{%
\subsection{Resampling for hyperparameter
tuning}\label{resampling-for-hyperparameter-tuning}}

CV is also widely used to assess model performance when tuning the
hyperparameters of flexible machine-learning models, and this is also
supported by the \pkg{mlr3} framework. Using the CV methods introduced
here, \pkg{mlr3} can therefore be used to optimized models to show an
improved performance in specific spatial or spatiotemporal predictive
setting \citep{schratz2019}. Such an optimization may, for example,
result in a reduced maximum tree depth or increased minimum node size in
the Ecuador case study, since these hyperparameter settings would result
in a stronger generalization and reduced overfitting.

We recommend using nested CV for this purpose. In nested CV, an
``inner'' CV is performed on each CV training set, since hyperparameter
tuning is an integral part of model fitting that should not be able to
use information from the CV test set. In such scenarios it is
recommended to use the same spatial resampling method for the inner CV
(hyperparameter tuning) as for the outer CV (model assessment) in order
to use the appropriate objective function for optimization. See
\citet{schratz2019} for more details as well as chapter 11 of
\href{https://geocompr.robinlovelace.net/spatial-cv.html}{Geocomputation
with R} \citep{lovelace2019}.

\hypertarget{additional-practical-issues}{%
\subsection{Additional practical
issues}\label{additional-practical-issues}}

Since \pkg{mlr3spatiotempcv} harvests already implemented resampling
methods from existing R packages, the broader overview presented in this
work has highlighted that there are still several gaps that may need to
be closed in the future, if specific use cases require those features.

For example, buffering, or the use of a spatial or temporal separation
distance between training and test sets, is currently only implemented
for some methods (\texttt{"spcv\_buffer"}, \texttt{"spcv\_disc"}, and
\texttt{"sptcv\_cstf"} with both \texttt{space\_var} and
\texttt{time\_var}). Its use should, however, be limited to use cases
involving a prediction distance, as a buffer zone reduces the size of
the training sample and introduces the risk of geographically biased
training data.

CV is often executed repeatedly to reduce the possible influence of
random variability on CV estimates. In general, only methods that
involve a random mechanism for generating or resampling blocks are
suited for this. In leave-one-block-out CV, coordinate-based and
environmental clustering (\texttt{"spcv\_coords"}, \texttt{"spcv\_env"}
and \texttt{"sptcv\_cluto"}) achieve this as their clusters are
generated based on random seeds. However, experience with
\texttt{"spcv\_coords"} shows that clusters from repeated executions may
in some situations be nearly identical to each other, resulting in very
little variability between CV repetitions. While this effect also
depends on the variable used for clustering, similar effects could
potentially also apply to \texttt{"spcv\_env"} and
\texttt{"sptcv\_cluto"} methods. However, such effects are more
difficult to quantify because selected features of these methods are
always different, in contrast to \texttt{"spcv\_coords"} which always
uses coordinates for clustering. This issue is even more critical in CV
at the block level with \texttt{"spcv\_block"} with options
\texttt{selection\ =\ "systematic"} and
\texttt{selection\ =\ "checkerboard"} because identical folds are
assigned in each repetition. In contrast, \texttt{"spcv\_block"} with
option \texttt{selection\ =\ "random"} avoids this problem.

\hypertarget{conclusion-and-outlook}{%
\section{Conclusion and outlook}\label{conclusion-and-outlook}}

The \pkg{mlr3spatiotempcv} package is the first package to bundle and
categorize spatiotemporal resampling methods implemented in multiple
other packages in \proglang{R}. The available resampling techniques
allow users to vary the scale or granularity of the resampled
spatiotemporal units as well as their shape and possible buffer distance
between training and test samples. These settings may account for the
specific characteristics of spatiotemporal prediction tasks, but
modelers now have to make the important decision of choosing a method
that is adequate for their situation. They are advised to focus on the
spatial or spatiotemporal structure of the model's prediction task,
consider the structure of the learning sample at hand, and think about
how the autocorrelation between training and test samples might affect
their model assessment and selection.

The compilation of resampling techniques in \pkg{mlr3spatiotempcv} is by
no means complete. Additional methods or parameters may therefore be
added in the future as they become available in upstream package or are
contributed directly to this package.

Spatiotemporal cross-validation as a paradigm is not yet fully
established in scientific workflows, although it has been discussed
intensively for more than a decade now. We anticipate that making the
existing methods easily accessible to users is an important step to
foster the acceptance of spatiotemporal cross-validation in the
community and to allow modelers to produce bias-reduced model
assessments in environmental and ecological studies.

\begin{CodeChunk}
\begin{CodeOutput}
R version 4.2.1 (2022-06-23)
Platform: aarch64-apple-darwin20 (64-bit)
Running under: macOS Ventura 13.0

Matrix products: default

locale:
[1] en_US.UTF-8/en_US.UTF-8/en_US.UTF-8/C/en_US.UTF-8/en_US.UTF-8

attached base packages:
[1] stats     graphics  grDevices utils     datasets  methods   base

other attached packages:
 [1] mlr3learners_0.5.4          patchwork_1.1.2             mlr3spatiotempcv_2.0.3.9000 mlr3_0.14.0-9000
 [5] pak_0.3.1.9000              browse_0.1.0                remotes_2.4.2               fledge_0.1.0.9019
 [9] reprex_2.0.2                teamtools_0.1.0             usethis_2.1.6

loaded via a namespace (and not attached):
  [1] paradox_0.10.0       colorspace_2.0-3     ellipsis_0.3.2       class_7.3-20         rgdal_1.5-32
  [6] rprojroot_2.0.3      markdown_1.3         fs_1.5.2             gridtext_0.1.5       ggtext_0.1.2
 [11] rstudioapi_0.14      proxy_0.4-27         listenv_0.8.0        farver_2.1.1         sperrorest_3.0.5
 [16] fansi_1.0.3          ranger_0.14.1        xml2_1.3.3           codetools_0.2-18     knitr_1.40
 [21] jsonlite_1.8.3       png_0.1-7            compiler_4.2.1       httr_1.4.4           backports_1.4.1
 [26] Matrix_1.5-1         assertthat_0.2.1     fastmap_1.1.0        lazyeval_0.2.2       cli_3.4.1
 [31] later_1.3.0          s2_1.1.0             htmltools_0.5.3      prettyunits_1.1.1    tools_4.2.1
 [36] gtable_0.3.1         glue_1.6.2           dplyr_1.0.10         wk_0.7.0             tinytex_0.42
 [41] Rcpp_1.0.9           raster_3.6-3         vctrs_0.5.0          crosstalk_1.2.0      xfun_0.34
 [46] stringr_1.4.1        mlr3measures_0.5.0   globals_0.16.1       ps_1.7.2             lifecycle_1.0.3
 [51] future_1.28.0        terra_1.6-17         scales_1.2.1         lgr_0.4.4            hms_1.1.2
 [56] parallel_4.2.2       rticles_0.24.2       yaml_2.3.6           curl_4.3.3           reticulate_1.26
 [61] ggplot2_3.3.6        blockCV_2.1.4        stringi_1.7.8        e1071_1.7-12         checkmate_2.1.0
 [66] palmerpenguins_0.1.1 pkgbuild_1.3.1       rlang_1.0.6          pkgconfig_2.0.3      commonmark_1.8.1
 [71] evaluate_0.17        lattice_0.20-45      ROCR_1.0-11          purrr_0.3.5          sf_1.0-8
 [76] htmlwidgets_1.5.4    processx_3.8.0       tidyselect_1.2.0     here_1.0.1           parallelly_1.32.1
 [81] magrittr_2.0.3       R6_2.5.1             generics_0.1.3       DBI_1.1.3            pillar_1.8.1
 [86] withr_2.5.0          units_0.8-0          sp_1.5-0             tibble_3.1.8         future.apply_1.9.1
 [91] crayon_1.5.2         uuid_1.1-0           KernSmooth_2.23-20   utf8_1.2.2           plotly_4.10.0
 [96] rmarkdown_2.17       progress_1.2.2       grid_4.2.2           data.table_1.14.4    callr_3.7.2
[101] git2r_0.30.1         mlr3misc_0.11.0      digest_0.6.30        classInt_0.4-8       tidyr_1.2.1
[106] munsell_0.5.0        viridisLite_0.4.1    rsthemes_0.4.0
\end{CodeOutput}
\end{CodeChunk}

\renewcommand\refname{References}
\bibliography{paper.bib}

\begin{thebibliography}{60}
\newcommand{\enquote}[1]{``#1''}
\providecommand{\natexlab}[1]{#1}
\providecommand{\url}[1]{\texttt{#1}}
\providecommand{\urlprefix}{URL }
\expandafter\ifx\csname urlstyle\endcsname\relax
  \providecommand{\doi}[1]{doi:\discretionary{}{}{}#1}\else
  \providecommand{\doi}{doi:\discretionary{}{}{}\begingroup
  \urlstyle{rm}\Url}\fi
\providecommand{\eprint}[2][]{\url{#2}}

\bibitem[{Anderson \emph{et~al.}(2005)Anderson, Turner, Forester, Zhu, Boyce,
  Beyer, and Stowell}]{anderson2005}
Anderson P, Turner MG, Forester JD, Zhu J, Boyce MS, Beyer H, Stowell L (2005).
\newblock \enquote{Scale-{{Dependent Summer Resource Selection}} by
  {{Reintroduced Elk}} in {{Wisconsin}}, {{USA}}.}
\newblock \emph{The Journal of Wildlife Management}, \textbf{69}(1), 298--310.
\newblock ISSN 0022-541X.
\newblock \urlprefix\url{https://www.jstor.org/stable/3803606}.

\bibitem[{Arlot and Celisse(2010)}]{arlot2010}
Arlot S, Celisse A (2010).
\newblock \enquote{A survey of cross-validation procedures for model
  selection.}
\newblock \emph{Statistics Surveys}, \textbf{4}(none), 40--79.
\newblock ISSN 1935-7516.
\newblock \doi{10.1214/09-SS054}.

\bibitem[{Bebber and Butt(2017)}]{bebber2017}
Bebber DP, Butt N (2017).
\newblock \enquote{Tropical protected areas reduced deforestation carbon
  emissions by one third from 2000\textendash 2012.}
\newblock \emph{Scientific Reports}, \textbf{7}(1), 14005.
\newblock ISSN 2045-2322.
\newblock \doi{10.1038/s41598-017-14467-w}.

\bibitem[{Becker \emph{et~al.}(2020)Becker, Binder, Bischl, Lang, Pfisterer,
  Reich, Richter, Schratz, and Sonabend}]{mlr3book}
Becker M, Binder M, Bischl B, Lang M, Pfisterer F, Reich NG, Richter J, Schratz
  P, Sonabend R (2020).
\newblock \emph{\{mlr3 book\}}.
\newblock \urlprefix\url{https://mlr3book.mlr-org.com}.

\bibitem[{Bengio and Grandvalet(2004)}]{bengio2004}
Bengio Y, Grandvalet Y (2004).
\newblock \enquote{No {{Unbiased Estimator}} of the {{Variance}} of {{K-Fold
  Cross-Validation}}.}
\newblock \emph{The Journal of Machine Learning Research}, \textbf{5},
  1089--1105.
\newblock ISSN 1532-4435.
\newblock
  \urlprefix\url{https://www.jmlr.org/papers/volume5/grandvalet04a/grandvalet04a.pdf}.

\bibitem[{Bergmeir \emph{et~al.}(2018)Bergmeir, Hyndman, and
  Koo}]{bergmeir2018}
Bergmeir C, Hyndman RJ, Koo B (2018).
\newblock \enquote{A note on the validity of cross-validation for evaluating
  autoregressive time series prediction.}
\newblock \emph{Computational Statistics \& Data Analysis}, \textbf{120},
  70--83.
\newblock ISSN 0167-9473.
\newblock \doi{10.1016/j.csda.2017.11.003}.

\bibitem[{Binder \emph{et~al.}(2021)Binder, Pfisterer, Lang, Schneider,
  Kotthoff, and Bischl}]{mlr3pipelines}
Binder M, Pfisterer F, Lang M, Schneider L, Kotthoff L, Bischl B (2021).
\newblock \enquote{{{mlr3pipelines}} - flexible machine learning pipelines in
  r.}
\newblock \emph{Journal of Machine Learning Research}, \textbf{22}(184), 1--7.
\newblock \urlprefix\url{https://jmlr.org/papers/v22/21-0281.html}.

\bibitem[{Brenning(2005)}]{brenning2005}
Brenning A (2005).
\newblock \enquote{Spatial prediction models for landslide hazards: review,
  comparison and evaluation.}
\newblock \emph{Natural Hazards and Earth System Sciences}, \textbf{5}(6),
  853--862.
\newblock ISSN 1561-8633.
\newblock \doi{10.5194/nhess-5-853-2005}.

\bibitem[{Brenning(2012)}]{brenning2012}
Brenning A (2012).
\newblock \enquote{Spatial cross-validation and bootstrap for the assessment of
  prediction rules in remote sensing: {{The R}} package sperrorest.}
\newblock In \emph{2012 {{IEEE}} international geoscience and remote sensing
  symposium}. {IEEE}.
\newblock \doi{10.1109/igarss.2012.6352393}.

\bibitem[{Brenning and Lausen(2008)}]{brenning2008}
Brenning A, Lausen B (2008).
\newblock \enquote{Estimating error rates in the classification of paired
  organs.}
\newblock \emph{Statistics in Medicine}, \textbf{27}(22), 4515--4531.
\newblock ISSN 0277-6715.
\newblock \doi{10.1002/sim.3310}.

\bibitem[{Brenning \emph{et~al.}(2015)Brenning, Schwinn, {Ruiz-P{\'a}ez}, and
  Muenchow}]{brenning2015}
Brenning A, Schwinn M, {Ruiz-P{\'a}ez} AP, Muenchow J (2015).
\newblock \enquote{Landslide susceptibility near highways is increased by 1
  order of magnitude in the {{Andes}} of southern {{Ecuador}}, {{Loja}}
  province.}
\newblock \emph{Natural Hazards and Earth System Sciences}, \textbf{15}(1),
  45--57.
\newblock ISSN 1561-8633.
\newblock \doi{10.5194/nhess-15-45-2015}.

\bibitem[{Cawley and Talbot(2010)}]{cawley2010}
Cawley GC, Talbot NLC (2010).
\newblock \enquote{On {{Over-fitting}} in {{Model Selection}} and {{Subsequent
  Selection Bias}} in {{Performance Evaluation}}.}
\newblock \emph{Journal of Machine Learning Research}, \textbf{11}(70),
  2079--2107.
\newblock ISSN 1533-7928.
\newblock \urlprefix\url{http://jmlr.org/papers/v11/cawley10a.html}.

\bibitem[{Cressie(1993)}]{cressie1993}
Cressie NAC (1993).
\newblock \emph{Statistics for {{Spatial Data}}}.
\newblock {John Wiley \& Sons}.
\newblock \doi{10.1002/9781119115151}.

\bibitem[{Diesing(2020)}]{diesing2020}
Diesing M (2020).
\newblock \enquote{Deep-sea sediments of the global ocean.}
\newblock \emph{Earth System Science Data}, \textbf{12}(4), 3367--3381.
\newblock ISSN 1866-3508.
\newblock \doi{10.5194/essd-12-3367-2020}.

\bibitem[{Efron and Gong(1983)}]{efron1983}
Efron B, Gong G (1983).
\newblock \enquote{A {{Leisurely Look}} at the {{Bootstrap}}, the
  {{Jackknife}}, and {{Cross-Validation}}.}
\newblock \emph{The American Statistician}, \textbf{37}(1), 36--48.
\newblock ISSN 0003-1305.
\newblock \doi{10.1080/00031305.1983.10483087}.

\bibitem[{Egli and H{\"o}pke(2020)}]{egli2020}
Egli S, H{\"o}pke M (2020).
\newblock \enquote{{{CNN-Based Tree Species Classification Using High
  Resolution RGB Image Data}} from {{Automated UAV Observations}}.}
\newblock \emph{Remote Sensing}, \textbf{12}(23), 3892.
\newblock \doi{10.3390/rs12233892}.

\bibitem[{Endicott \emph{et~al.}(2017)Endicott, Drescher, and
  Brenning}]{endicott2017}
Endicott S, Drescher M, Brenning A (2017).
\newblock \enquote{Modelling the spread of {{European}} buckthorn in the
  {{Region}} of {{Waterloo}}.}
\newblock \emph{Biological Invasions}, \textbf{19}(10), 2993--3011.
\newblock ISSN 1573-1464.
\newblock \doi{10.1007/s10530-017-1504-3}.

\bibitem[{Escobar \emph{et~al.}(2021)Escobar, Helmstetter, Jarvie,
  Mont{\'u}far, Balslev, and Couvreur}]{escobar2021}
Escobar S, Helmstetter AJ, Jarvie S, Mont{\'u}far R, Balslev H, Couvreur TLP
  (2021).
\newblock \enquote{Pleistocene climatic fluctuations promoted alternative
  evolutionary histories in {{Phytelephas}} aequatorialis, an endemic palm from
  western {{Ecuador}}.}
\newblock \emph{Journal of Biogeography}, \textbf{48}(5), 1023--1037.
\newblock ISSN 1365-2699.
\newblock \doi{10.1111/jbi.14055}.

\bibitem[{Gao \emph{et~al.}(2019)Gao, Liang, Yin, Ge, Feng, Wu, Hou, Liu, and
  Xie}]{gao2019}
Gao J, Liang T, Yin J, Ge J, Feng Q, Wu C, Hou M, Liu J, Xie H (2019).
\newblock \enquote{Estimation of {{Alpine Grassland Forage Nitrogen Coupled}}
  with {{Hyperspectral Characteristics}} during {{Different Growth Periods}} on
  the {{Tibetan Plateau}}.}
\newblock \emph{Remote Sensing}, \textbf{11}(18), 2085.
\newblock \doi{10.3390/rs11182085}.

\bibitem[{Gei{\ss} \emph{et~al.}(2017)Gei{\ss}, Aravena~Pelizari, Schrade,
  Brenning, and Taubenb{\"o}ck}]{geiss2017}
Gei{\ss} C, Aravena~Pelizari P, Schrade H, Brenning A, Taubenb{\"o}ck H (2017).
\newblock \enquote{On the {{Effect}} of {{Spatially Non-Disjoint Training}} and
  {{Test Samples}} on {{Estimated Model Generalization Capabilities}} in
  {{Supervised Classification With Spatial Features}}.}
\newblock \emph{IEEE Geoscience and Remote Sensing Letters}, \textbf{14}(11),
  2008--2012.
\newblock ISSN 1558-0571.
\newblock \doi{10.1109/LGRS.2017.2747222}.

\bibitem[{Hand(1997)}]{hand1997}
Hand D (1997).
\newblock \emph{Construction and {{Assessment}} of {{Classification Rules}}}.
\newblock {Wiley}, {New York}.
\newblock
  \urlprefix\url{https://www.wiley.com/en-sg/Construction+and+Assessment+of+Classification+Rules-p-9780471965831}.

\bibitem[{Hartigan and Wong(1979)}]{hartigan1979a}
Hartigan JA, Wong MA (1979).
\newblock \enquote{Algorithm {{AS}} 136: {{A K-Means Clustering Algorithm}}.}
\newblock \emph{Journal of the Royal Statistical Society C}, \textbf{28}(1),
  100--108.
\newblock ISSN 0035-9254.
\newblock \doi{10.2307/2346830}.

\bibitem[{Hijmans \emph{et~al.}(2020)Hijmans, Phillips, Leathwick, and
  Elith}]{hijmans2020}
Hijmans RJ, Phillips S, Leathwick J, Elith J (2020).
\newblock \emph{dismo: {{Species}} distribution modeling}.
\newblock \urlprefix\url{https://CRAN.R-project.org/package=dismo}.

\bibitem[{Hornik \emph{et~al.}(2012)Hornik, Feinerer, Kober, and
  Buchta}]{skmeans}
Hornik K, Feinerer I, Kober M, Buchta C (2012).
\newblock \enquote{Spherical {{k}}-{{Means}} clustering.}
\newblock \emph{Journal of Statistical Software}, \textbf{50}(10), 1--22.
\newblock \doi{10.18637/jss.v050.i10}.

\bibitem[{Hyndman and Koehler(2006)}]{hyndman2006}
Hyndman RJ, Koehler AB (2006).
\newblock \enquote{Another look at measures of forecast accuracy -
  {{ScienceDirect}}.}
\newblock \emph{International Journal of Forecasting}, \textbf{22}(4),
  679--688.
\newblock \doi{10.1016/j.ijforecast.2006.03.001}.

\bibitem[{Jensen \emph{et~al.}(2021)Jensen, Rao, Zhang, Gr{\o}n, Tian, Ma, and
  Svenning}]{jensen2021}
Jensen DA, Rao M, Zhang J, Gr{\o}n M, Tian S, Ma K, Svenning JC (2021).
\newblock \enquote{The potential for using rare, native species in
  reforestation\textendash{} {{A}} case study of yews ({{Taxaceae}}) in
  {{China}}.}
\newblock \emph{Forest Ecology and Management}, \textbf{482}, 118816.
\newblock ISSN 0378-1127.
\newblock \doi{10.1016/j.foreco.2020.118816}.

\bibitem[{Karasiak \emph{et~al.}(2021)Karasiak, Dejoux, Monteil, and
  Sheeren}]{karasiak2021}
Karasiak N, Dejoux JF, Monteil C, Sheeren D (2021).
\newblock \enquote{Spatial dependence between training and test sets: another
  pitfall of classification accuracy assessment in remote sensing.}
\newblock \emph{Machine Learning}.
\newblock ISSN 1573-0565.
\newblock \doi{10.1007/s10994-021-05972-1}.

\bibitem[{Kasurak \emph{et~al.}(2011)Kasurak, Kelly, and
  Brenning}]{kasurak2011}
Kasurak A, Kelly R, Brenning A (2011).
\newblock \enquote{Linear mixed modelling of snow distribution in the central
  {{Yukon}}.}
\newblock \emph{Hydrological Processes}, \textbf{25}(21), 3332--3346.
\newblock ISSN 1099-1085.
\newblock \doi{10.1002/hyp.8168}.

\bibitem[{Kuhn and Wickham(2020)}]{kuhn2020}
Kuhn M, Wickham H (2020).
\newblock \emph{Tidymodels: a collection of packages for modeling and machine
  learning using tidyverse principles.}
\newblock \urlprefix\url{https://www.tidymodels.org}.

\bibitem[{Lang \emph{et~al.}(2020)Lang, Au, Coors, and Schratz}]{mlr3learners}
Lang M, Au Q, Coors S, Schratz P (2020).
\newblock \emph{mlr3learners: {{Recommended Learners}} for 'mlr3'}.
\newblock \urlprefix\url{https://CRAN.R-project.org/package=mlr3learners}.

\bibitem[{Lang \emph{et~al.}(2019)Lang, Binder, Richter, Schratz, Pfisterer,
  Coors, Au, Casalicchio, Kotthoff, and Bischl}]{mlr3}
Lang M, Binder M, Richter J, Schratz P, Pfisterer F, Coors S, Au Q, Casalicchio
  G, Kotthoff L, Bischl B (2019).
\newblock \enquote{{{mlr3}}: {{A}} modern object-oriented machine learning
  framework in {{R}}.}
\newblock \emph{Journal of Open Source Software}.
\newblock \doi{10.21105/joss.01903}.

\bibitem[{Lovelace \emph{et~al.}(2019)Lovelace, Nowosad, and
  Muenchow}]{lovelace2019}
Lovelace R, Nowosad J, Muenchow J (2019).
\newblock \emph{Geocomputation with {{R}}}.
\newblock {CRC Press}.
\newblock \urlprefix\url{https://geocompr.robinlovelace.net/}.

\bibitem[{Martin \emph{et~al.}(2008)Martin, Plourde, Ollinger, Smith, and
  McNeil}]{martin2008}
Martin ME, Plourde LC, Ollinger SV, Smith ML, McNeil BE (2008).
\newblock \enquote{A generalizable method for remote sensing of canopy nitrogen
  across a wide range of forest ecosystems.}
\newblock \emph{Remote Sensing of Environment}, \textbf{112}(9), 3511--3519.
\newblock ISSN 0034-4257.
\newblock \doi{10.1016/j.rse.2008.04.008}.

\bibitem[{Meyer(2020)}]{cast}
Meyer H (2020).
\newblock \emph{{{CAST}}: 'caret' applications for spatial-temporal models}.
\newblock \urlprefix\url{https://CRAN.R-project.org/package=CAST}.

\bibitem[{Meyer \emph{et~al.}(2018)Meyer, Reudenbach, Hengl, Katurji, and
  Nauss}]{meyer2018}
Meyer H, Reudenbach C, Hengl T, Katurji M, Nauss T (2018).
\newblock \enquote{Improving performance of spatio-temporal machine learning
  models using forward feature selection and target-oriented validation.}
\newblock \emph{Environmental Modelling \& Software}, \textbf{101}, 1--9.
\newblock ISSN 1364-8152.
\newblock \doi{10.1016/j.envsoft.2017.12.001}.

\bibitem[{M{\o}ller \emph{et~al.}(2021)M{\o}ller, Mulder, Heuvelink, Jacobsen,
  and Greve}]{moller2021}
M{\o}ller AB, Mulder VL, Heuvelink GBM, Jacobsen NM, Greve MH (2021).
\newblock \enquote{Can {{We Use Machine Learning}} for {{Agricultural Land
  Suitability Assessment}}?}
\newblock \emph{Agronomy}, \textbf{11}(4), 703.
\newblock \doi{10.3390/agronomy11040703}.

\bibitem[{Morera \emph{et~al.}(2021)Morera, {Mart{\'i}nez de Arag{\'o}n},
  Bonet, Liang, and {de-Miguel}}]{morera2021}
Morera A, {Mart{\'i}nez de Arag{\'o}n} J, Bonet JA, Liang J, {de-Miguel} S
  (2021).
\newblock \enquote{Performance of statistical and machine learning-based
  methods for predicting biogeographical patterns of fungal productivity in
  forest ecosystems.}
\newblock \emph{Forest Ecosystems}, \textbf{8}(1), 21.
\newblock ISSN 2197-5620.
\newblock \doi{10.1186/s40663-021-00297-w}.

\bibitem[{Muenchow \emph{et~al.}(2012)Muenchow, Brenning, and
  Richter}]{muenchow2012}
Muenchow J, Brenning A, Richter M (2012).
\newblock \enquote{Geomorphic process rates of landslides along a humidity
  gradient in the tropical {{Andes}}.}
\newblock \emph{Geomorphology}, \textbf{139--140}, 271--284.
\newblock ISSN 0169-555X.
\newblock \doi{10.1016/j.geomorph.2011.10.029}.

\bibitem[{Muscarella \emph{et~al.}(2014)Muscarella, Galante, {Soley-Guardia},
  Boria, Kass, Uriarte, and Anderson}]{muscarella2014}
Muscarella R, Galante PJ, {Soley-Guardia} M, Boria RA, Kass JM, Uriarte M,
  Anderson RP (2014).
\newblock \enquote{{{ENMeval}}: {{An R}} package for conducting spatially
  independent evaluations and estimating optimal model complexity for
  {{Maxent}} ecological niche models.}
\newblock \emph{Methods in Ecology and Evolution}, \textbf{5}(11), 1198--1205.
\newblock ISSN 2041-210X.
\newblock \doi{10.1111/2041-210X.12261}.

\bibitem[{Pebesma(2018)}]{pebesma2018}
Pebesma E (2018).
\newblock \enquote{Simple {{Features}} for {{R}}: {{Standardized Support}} for
  {{Spatial Vector Data}}.}
\newblock \emph{The R Journal}, \textbf{10}(1), 439--446.
\newblock ISSN 2073-4859.
\newblock
  \urlprefix\url{https://journal.r-project.org/archive/2018/RJ-2018-009/index.html}.

\bibitem[{Pedregosa \emph{et~al.}(2011)Pedregosa, Varoquaux, Gramfort, Michel,
  Thirion, Grisel, Blondel, Prettenhofer, Weiss, Dubourg, Vanderplas, Passos,
  Cournapeau, Brucher, Perrot, and Duchesnay}]{pedregosa2011}
Pedregosa F, Varoquaux G, Gramfort A, Michel V, Thirion B, Grisel O, Blondel M,
  Prettenhofer P, Weiss R, Dubourg V, Vanderplas J, Passos A, Cournapeau D,
  Brucher M, Perrot M, Duchesnay {\'E} (2011).
\newblock \enquote{Scikit-learn: {{Machine Learning}} in {{Python}}.}
\newblock \emph{Journal of Machine Learning Research}, \textbf{12}(85),
  2825--2830.
\newblock \urlprefix\url{http://jmlr.org/papers/v12/pedregosa11a.html}.

\bibitem[{Pe{\~n}a and Brenning(2015)}]{pena2015}
Pe{\~n}a M, Brenning A (2015).
\newblock \enquote{Assessing fruit-tree crop classification from {{Landsat-8}}
  time series for the {{Maipo Valley}}, {{Chile}}.}
\newblock \emph{Remote Sensing of Environment}, \textbf{171}, 234--244.
\newblock \doi{10.1016/j.rse.2015.10.029}.

\bibitem[{Ploton \emph{et~al.}(2020)Ploton, Mortier, {R{\'e}jou-M{\'e}chain},
  Barbier, Picard, Rossi, Dormann, Cornu, Viennois, Bayol, Lyapustin,
  {Gourlet-Fleury}, and P{\'e}lissier}]{ploton2020}
Ploton P, Mortier F, {R{\'e}jou-M{\'e}chain} M, Barbier N, Picard N, Rossi V,
  Dormann C, Cornu G, Viennois G, Bayol N, Lyapustin A, {Gourlet-Fleury} S,
  P{\'e}lissier R (2020).
\newblock \enquote{Spatial validation reveals poor predictive performance of
  large-scale ecological mapping models.}
\newblock \emph{Nature Communications}, \textbf{11}(1), 4540.
\newblock ISSN 2041-1723.
\newblock \doi{10.1038/s41467-020-18321-y}.

\bibitem[{Pohjankukka \emph{et~al.}(2017)Pohjankukka, Pahikkala, Nevalainen,
  and Heikkonen}]{pohjankukka2017}
Pohjankukka J, Pahikkala T, Nevalainen P, Heikkonen J (2017).
\newblock \enquote{Estimating the prediction performance of spatial models via
  spatial k-fold cross validation.}
\newblock \emph{International Journal of Geographical Information Science},
  \textbf{31}(10), 2001--2019.
\newblock \doi{10.1080/13658816.2017.1346255}.

\bibitem[{Reitz \emph{et~al.}(2021)Reitz, Graf, Schmidt, Ketzler, and
  Leuchner}]{reitz2021}
Reitz O, Graf A, Schmidt M, Ketzler G, Leuchner M (2021).
\newblock \enquote{Upscaling {{Net Ecosystem Exchange Over Heterogeneous
  Landscapes With Machine Learning}}.}
\newblock \emph{Journal of Geophysical Research: Biogeosciences},
  \textbf{126}(2), e2020JG005814.
\newblock ISSN 2169-8961.
\newblock \doi{10.1029/2020JG005814}.

\bibitem[{Rest \emph{et~al.}(2014)Rest, Pinaud, Monestiez, Chadoeuf, and
  Bretagnolle}]{rest2014}
Rest KL, Pinaud D, Monestiez P, Chadoeuf J, Bretagnolle V (2014).
\newblock \enquote{Spatial leave-one-out cross-validation for variable
  selection in the presence of spatial autocorrelation.}
\newblock \emph{Global Ecology and Biogeography}, \textbf{23}(7), 811--820.
\newblock ISSN 1466-8238.
\newblock \doi{10.1111/geb.12161}.

\bibitem[{Roberts \emph{et~al.}(2017)Roberts, Bahn, Ciuti, Boyce, Elith,
  {Guillera-Arroita}, Hauenstein, {Lahoz-Monfort}, Schr{\"o}der, Thuiller,
  Warton, Wintle, Hartig, and Dormann}]{roberts2017}
Roberts DR, Bahn V, Ciuti S, Boyce MS, Elith J, {Guillera-Arroita} G,
  Hauenstein S, {Lahoz-Monfort} JJ, Schr{\"o}der B, Thuiller W, Warton DI,
  Wintle BA, Hartig F, Dormann CF (2017).
\newblock \enquote{Cross-validation strategies for data with temporal, spatial,
  hierarchical, or phylogenetic structure.}
\newblock \emph{Ecography}, \textbf{40}(8), 913--929.
\newblock \doi{10.1111/ecog.02881}.

\bibitem[{Ru{\ss} and Brenning(2010)}]{russ2010}
Ru{\ss} G, Brenning A (2010).
\newblock \enquote{Data {{Mining}} in {{Precision Agriculture}}: {{Management}}
  of {{Spatial Information}}.}
\newblock In E~H{\"u}llermeier, R~Kruse, F~Hoffmann (eds.), \emph{Computational
  {{Intelligence}} for {{Knowledge-Based Systems Design}}}, Lecture {{Notes}}
  in {{Computer Science}}, pp. 350--359. {Springer-Verlag}, {Berlin,
  Heidelberg}.
\newblock ISBN 978-3-642-14049-5.
\newblock \doi{10.1007/978-3-642-14049-5_36}.

\bibitem[{Schratz \emph{et~al.}(2019)Schratz, Muenchow, Iturritxa, Richter, and
  Brenning}]{schratz2019}
Schratz P, Muenchow J, Iturritxa E, Richter J, Brenning A (2019).
\newblock \enquote{Hyperparameter tuning and performance assessment of
  statistical and machine-learning algorithms using spatial data.}
\newblock \emph{Ecological Modelling}, \textbf{406}, 109--120.
\newblock \doi{10.1016/j.ecolmodel.2019.06.002}.

\bibitem[{Sievert(2020)}]{plotly}
Sievert C (2020).
\newblock \emph{Interactive web-based data visualization with r, plotly, and
  shiny}.
\newblock {Chapman and Hall/CRC}.
\newblock ISBN 978-1-138-33145-7.
\newblock \urlprefix\url{https://plotly-r.com}.

\bibitem[{Stewart \emph{et~al.}(2021)Stewart, Elith, Fedrigo, Kasel, Roxburgh,
  Bennett, Chick, Fairman, Leonard, Kohout, Cripps, Durkin, and
  Nitschke}]{stewart2021}
Stewart SB, Elith J, Fedrigo M, Kasel S, Roxburgh SH, Bennett LT, Chick M,
  Fairman T, Leonard S, Kohout M, Cripps JK, Durkin L, Nitschke CR (2021).
\newblock \enquote{Climate extreme variables generated using monthly
  time-series data improve predicted distributions of plant species.}
\newblock \emph{Ecography}, \textbf{44}(4), 626--639.
\newblock ISSN 1600-0587.
\newblock \doi{10.1111/ecog.05253}.

\bibitem[{Thompson(2012)}]{thompson2012}
Thompson SK (2012).
\newblock \enquote{Sampling, {{Third Edition}}.}
\newblock In \emph{Sampling}, pp. i--xxi. {John Wiley \& Sons}.
\newblock ISBN 978-1-118-16293-4.
\newblock \doi{10.1002/9781118162934.fmatter}.

\bibitem[{Valavi \emph{et~al.}(2019)Valavi, Elith, {Lahoz-Monfort},
  {Guillera-Arroita}, Valavi, Elith, {Lahoz-Monfort}, and
  {Guillera-Arroita}}]{blockCV}
Valavi R, Elith J, {Lahoz-Monfort} JJ, {Guillera-Arroita} G, Valavi R, Elith J,
  {Lahoz-Monfort} JJ, {Guillera-Arroita} G (2019).
\newblock \enquote{{{blockCV}}: {{An R}} package for generating spatially or
  environmentally separated folds for k-fold cross-validation of species
  distribution models.}
\newblock \emph{Methods in Ecology and Evolution}, \textbf{10}(2), 225--232.
\newblock \doi{10.1111/2041-210X.13107}.

\bibitem[{Vanwinckelen and Blockeel(2012)}]{vanwinckelen2012}
Vanwinckelen G, Blockeel H (2012).
\newblock \enquote{On estimating model accuracy with repeated
  cross-validation.}
\newblock In \emph{{{BeneLearn}} 2012: {{Proceedings}} of the 21st
  {{Belgian-Dutch Conference}} on {{Machine Learning}}}, pp. 39--44.
\newblock ISBN 978-94-6197-044-2.
\newblock \urlprefix\url{https://lirias.kuleuven.be/1655861}.

\bibitem[{Wickham(2016)}]{ggplot2}
Wickham H (2016).
\newblock \emph{ggplot2: {{Elegant}} graphics for data analysis}.
\newblock {Springer-Verlag New York}.
\newblock ISBN 978-3-319-24277-4.
\newblock \urlprefix\url{https://ggplot2.tidyverse.org}.

\bibitem[{Willmott and Matsuura(2006)}]{willmott2006}
Willmott CJ, Matsuura K (2006).
\newblock \enquote{On the use of dimensioned measures of error to evaluate the
  performance of spatial interpolators.}
\newblock \emph{International Journal of Geographical Information Science},
  \textbf{20}(1), 89--102.
\newblock ISSN 1365-8816.
\newblock \doi{10.1080/13658810500286976}.

\bibitem[{Wright and Ziegler(2017)}]{ranger}
Wright MN, Ziegler A (2017).
\newblock \enquote{ranger: {{A Fast Implementation}} of {{Random Forests}} for
  {{High Dimensional Data}} in {{C}}++ and {{R}}.}
\newblock \emph{Journal of Statistical Software}, \textbf{77}(1), 1--17.
\newblock \doi{10.18637/jss.v077.i01}.

\bibitem[{Wu \emph{et~al.}(2020)Wu, Luo, Dong, Gao, Hu, Wu, Sun, and
  Liu}]{wu2020}
Wu T, Luo J, Dong W, Gao L, Hu X, Wu Z, Sun Y, Liu J (2020).
\newblock \enquote{Disaggregating {{County-Level Census Data}} for {{Population
  Mapping Using Residential Geo-Objects With Multisource Geo-Spatial Data}}.}
\newblock \emph{IEEE Journal of Selected Topics in Applied Earth Observations
  and Remote Sensing}, \textbf{13}, 1189--1205.
\newblock ISSN 2151-1535.
\newblock \doi{10.1109/JSTARS.2020.2974896}.

\bibitem[{Zhao and Karypis(2002)}]{cluto}
Zhao Y, Karypis G (2002).
\newblock \enquote{Evaluation of hierarchical clustering algorithms for
  document datasets.}
\newblock In \emph{Proceedings of the eleventh international conference on
  {{Information}} and knowledge management}, pp. 515--524.
\newblock \doi{10.1145/584792.584877}.

\bibitem[{Zurell \emph{et~al.}(2020)Zurell, Zimmermann, Gross, Baltensweiler,
  Sattler, and W{\"u}est}]{zurell2020}
Zurell D, Zimmermann NE, Gross H, Baltensweiler A, Sattler T, W{\"u}est RO
  (2020).
\newblock \enquote{Testing species assemblage predictions from stacked and
  joint species distribution models.}
\newblock \emph{Journal of Biogeography}, \textbf{47}(1), 101--113.
\newblock ISSN 1365-2699.
\newblock \doi{10.1111/jbi.13608}.

\end{thebibliography}

\end{document}